\documentclass[conference]{IEEEtran}
\IEEEoverridecommandlockouts
\usepackage{cite}
\usepackage{amsmath,amssymb,amsfonts}
\usepackage{algorithmic}
\usepackage{graphicx}
\usepackage{textcomp}
\usepackage{xcolor}
\usepackage{stfloats}

\usepackage{amsmath}
\usepackage{mathtools}
\usepackage{amssymb}
\usepackage{float}
\usepackage{bm}
\usepackage{mathdots}
\usepackage{subfigure}
\usepackage{hyperref}
\usepackage{enumitem}   
\usepackage{balance}

\def\BibTeX{{\rm B\kern-.05em{\sc i\kern-.025em b}\kern-.08em
    T\kern-.1667em\lower.7ex\hbox{E}\kern-.125emX}}

\newcommand\aug{\fboxsep=-\fboxrule\!\!\!\fbox{\strut}\!\!\!}

\begin{document}

\title{Stable Modular Control via Contraction Theory \\ for Reinforcement Learning}

\author{\IEEEauthorblockN{ Bing Song}
\IEEEauthorblockA{\textit{HP-NTU Digital}\\ \textit{Manufacturing Corporate Lab} \\
Singapore,  Singapore 639798 \\
bing.song@ntu.edu.sg}
\and
\IEEEauthorblockN{Jean-Jacques Slotine}
\IEEEauthorblockA{\textit{Nonlinear Systems Laboratory} \\
\textit{Massachusetts Institute of Technology}\\
Cambridge MA 02139-4307   \\
jjs@mit.edu}
\and
\IEEEauthorblockN{Quang-Cuong Pham}
\IEEEauthorblockA{\textit{ HP-NTU Digital}\\ \textit{Manufacturing Corporate Lab} \\
Singapore,  Singapore 639798 \\
cuong@ntu.edu.sg}
}

\maketitle

\newcommand{\TODO}[1]{{\color{red} {\bf #1}}}

\begin{abstract}
We propose a novel way to integrate control techniques with reinforcement learning (RL) for stability, robustness, and generalization: leveraging contraction theory to realize modularity in neural control, which ensures that combining stable subsystems can automatically preserve the stability. We realize such modularity via signal composition and dynamic decomposition. Signal composition creates the latent space, within which RL applies to maximizing rewards. Dynamic decomposition is realized by coordinate transformation that creates an auxiliary space, within which the latent signals are coupled in the way that their combination can preserve stability provided each signal, that is, each subsystem, has stable self-feedbacks. Leveraging modularity, the nonlinear stability problem is deconstructed into algebraically solvable ones, the stability of the subsystems in the auxiliary space, yielding linear constraints on the input gradients of control networks that can be as simple as switching the signs of network weights. This minimally invasive method for stability allows arguably easy integration into the modular neural architectures in machine learning, like hierarchical RL, and improves their performance. We demonstrate in simulation the necessity and the effectiveness of our method: the necessity for robustness and generalization, and the effectiveness in improving hierarchical RL for manipulation learning. 
\end{abstract}

\section{Introduction}


Reinforcement learning (RL) has been struggling with control stability, robustness, and generalization. How to leverage control theoretical results to solve those issues is one fundamental question. Currently, control theoretical results are employed in RL to create some priors\cite{cheng2019control}, to design loss functions\cite{berkenkamp2017safe}, or to decrease the gap between the simulated environments and the real-world applications by improving dynamic models\cite{singh2018learning}, producing adversarial samples\cite{mandlekar2017adversarially}, etc. 

Here we propose a novel way to integrate control with RL: applying contraction theory\cite{lohmiller1998contraction} to creating modular architectures in neural control. Leveraging modularity, we develop a minimally invasive way to realize stable control with RL. 

\begin{figure}[t!]
            \centering
   \includegraphics[width=2.5in]{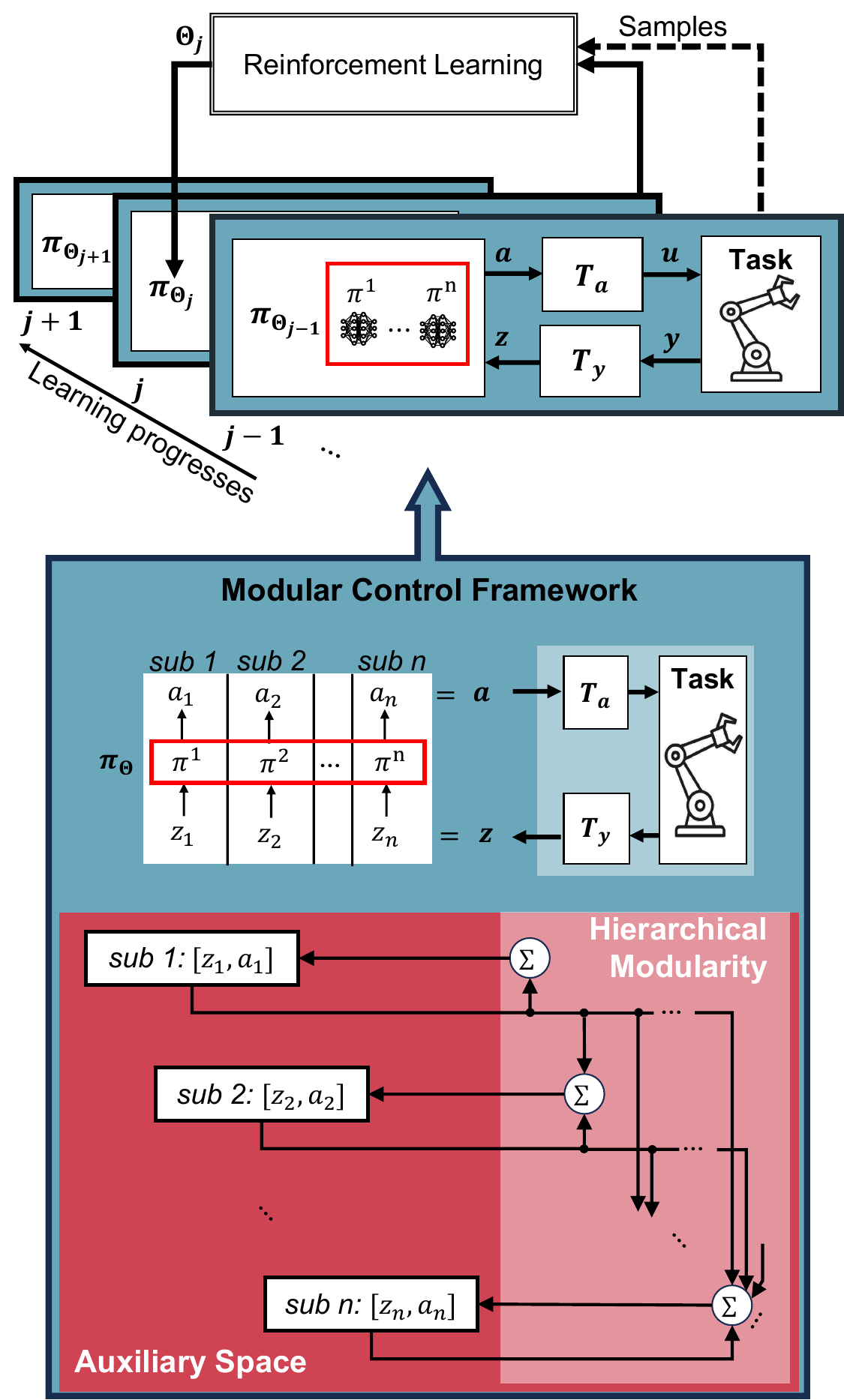}
    \caption{Modular control framework for RL. The modularity, for example, hierarchical combination of subsystems, is realized in the auxiliary space by coordinate transformation ($T_y, T_a$). Stability can be automatically guaranteed provided each subsystem has stable self-feedbacks, yielding linear constraints on the input gradients of control networks that can be as simple as switching the signs of weights.}\label{fig:model}  
   \end{figure}
   
To guarantee stability is more challenging for RL than for nonlinear control. 
RL targets at control problems likely involving large variations, uncertainties, unknowns, and non-convexities. To handle those difficulties, RL characterizes physical systems with distributions (rigorously, transition functions) and optimizes neural control according to estimated values (e.g., value functions and gradients). However, the stability of control is mathematically defined in nonlinear system theory for the underlying dynamics behind those distributions and estimations.

Complexities arise from this mismatch in formulation: stabilizing control of physical dynamics based on models vs. optimizing control for task distributions based on data. 

On one hand, it raises unique concerns. For example, in \cite{song2022stability}, we show that local optimal policies can contain unstable control for some dynamic parameters in the task distribution, even when there exist better policies with higher expected returns that provide stable control for all parameters. Overfitting to such instabilities can deteriorate generalization. 

On the other hand, it creates intrinsically difficult questions to extend control-theoretical results to RL. For example, the wide success of neural certificates in nonlinear control\cite{dawson2022safe}, both empirically and theoretically, has not yet been equally transferred to RL. The success of neural certificates depends on that as more stable trajectories are collected to learn the certificates, the bounds on the generalization error of the data-driven certificates to new trajectories are shrinking\cite{boffi2021learning}. The controllers are derived or optimized based on the learned certificates to perform stable control. Extending neural certificates to RL, the controllers need to be adapted simultaneously to maximize the estimated value functions for possibly non-convex problems and for possibly much larger variations in dynamic parameters. It raises mathematically challenging questions like whether the generalization error bounds for the certificates are still shrinking and whether the exploration for value estimation is unnecessarily limited. Currently in safe RL, there exists intuitive methods like weighted sum of objectives \cite{qin2021learning} and alternative updates\cite{luo2021learning}, and complicated methods like unifying the objective functions into a multi-timescale min-max-min optimization problem\cite{ma2022joint}. All lack the analysis on the interference between the generalization bounds of the certificates and the estimation of value functions. Expertise in both nonlinear control and RL is likely required for the empirical certificate loss design as well as task-specific hyperparameter tuning. 

In contrast, learning motor skills with stable control seems easy and natural to animals. It is believed in both machine learning\cite{reed2022generalist} and nonlinear control\cite{slotine2001modularity} that modularity is the key. Here we introduce into RL the modularity from nonlinear control, in particular, from contraction theory\cite{lohmiller1998contraction}.

More than functional specialization, modularity in biology also ensures that combining stable subsystems can automatically preserve the stability, which can be mathematically characterized by contraction theory\cite{slotine2003modular}. 

We leverage such modularity to deconstruct the nonlinear stability problem into algebraically solvable ones, the stability of elementary subsystems, yielding linear constraints on the input gradients of control networks (network Jacobians). 

\begin{figure}[t!]
    \centering 
        \includegraphics[width=3.5in]{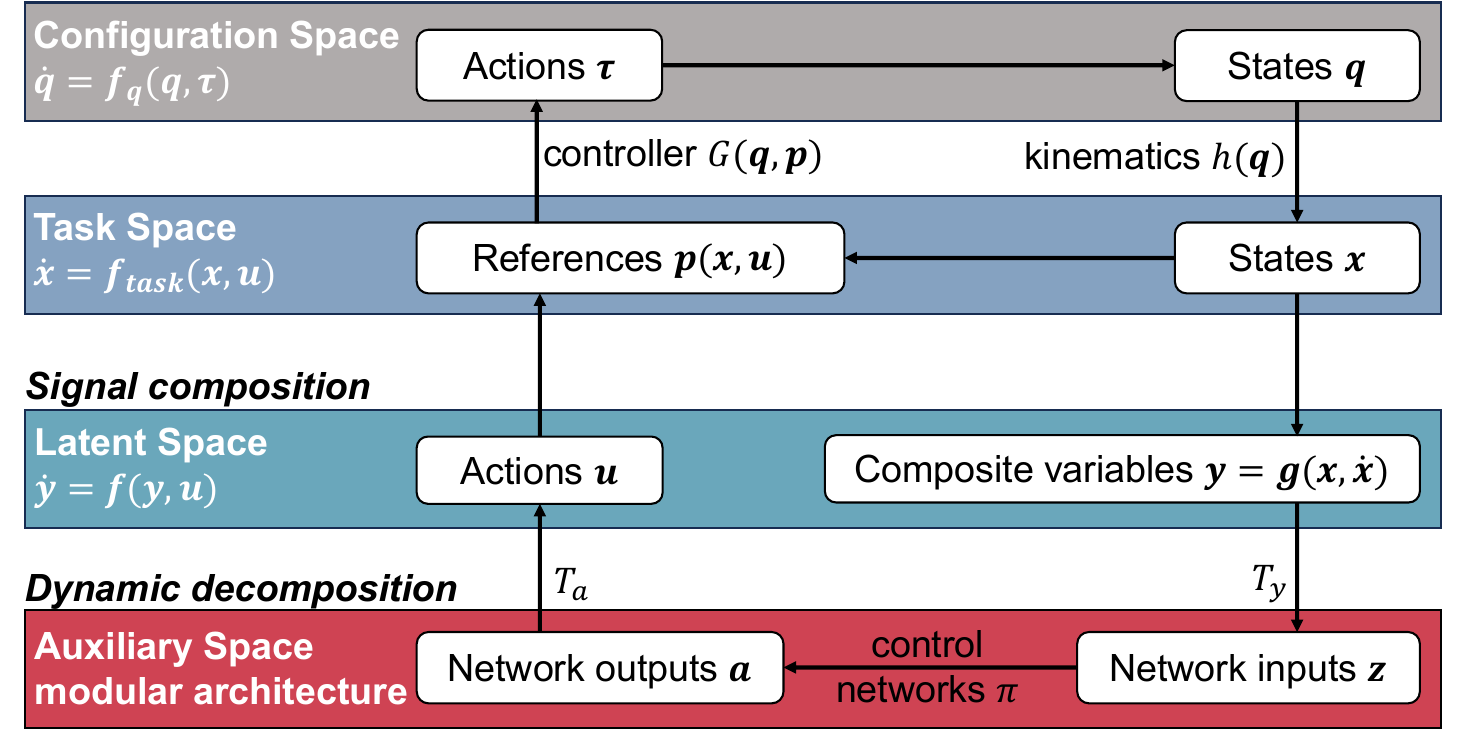}
    \caption{Signal flows across different spaces. Signal composition creates the latent space, within which neural control with RL applies to maximizing rewards. Provided that the composition of latent signals is a contracting system, the stability of latent signals implies the stability of physical dynamics. Dynamic decomposition in the auxiliary space further simplifies the stability of latent signals into algebraically solvable ones via modularity.} 
    \label{fig:composition}
\end{figure}

Such modularity is realized by signal composition and dynamic decomposition. Signal composition creates the latent space, within which RL applies to maximizing rewards. Dynamic decomposition is realized by coordinate transformation that creates an auxiliary space, within which the latent signals are coupled in the way that their combination can preserve stability as long as each signal has stable self-feedbacks, for example, the hierarchical type in Fig.~\ref{fig:model}. The graph of the signal flows across different spaces is illustrated in Fig.~\ref{fig:composition}. 

The modular architecture simplifies the nonlinear stability problem. Provided stable elementary subsystems in the auxiliary space, the latent signals in the latent space are stable, converging with bounded transients. Provided that the composition of latent signals is also a contracting system, the convergence of latent signals implies the stability of physical dynamics. 

In this paper, we use control techniques to create latent signals, in particular, composite variables that mix errors in position and velocity, which is common in biological nervous systems\cite{slotine2001modularity}. This framework may be extended to latent signals from machine learning.

To decompose dynamics in the auxiliary space, the dynamic model in the latent space is required, specifically, the partial derivatives. For unknown environments, one may apply system identification, learning to predict those partial derivatives. Here we provide another solution, leveraging task space controllers\cite{nakanishi2008operational} to build a latent-space model composed of the controller gains. Robot models are still required.

Robustness of our method to model errors and bounded disturbances can be characterized via contraction theory.

Our method provides a minimally invasive way for stable RL control. The pseudocode of RL algorithms remains the same, while the graph of control policies should be adapted into the modular control framework with coordinate transformation and the constrained control networks, as well as adding function compositions that define the RL states (latent signals) and integrating RL actions with existing controllers. The stability constraints can be as simple as switching the signs of network weights for multilayer perceptrons (MLPs).



Following the implementation and limitations, we demonstrate the necessity and the effectiveness of our method in simulation. The necessity for robustness and generalization is illustrated in the complementary paper \cite{song2022stability} and we briefly include the results here. The effectiveness is demonstrated with hierarchical RL (two TD3\cite{fujimoto2018addressing} agents, one for planning and the other for control), learning two manipulation scenarios: peg push and peg maze. The stability constraints on low-level control guarantee consistent dynamic behaviors against perturbations and variations in initial distributions, mitigating concerns like non-stationary distributions. Results shows that our method outperforms (i) the state-of-the-art data-efficient hierarchical RL\cite{nachum2018data}, (ii) hierarchical RL with unconstrained modular control, and (iii) one-level RL with stable non-neural controllers.

\subsection{Technical Contributions.} Our main contribution is the method that builds an auxiliary space to decompose the dynamics for modular neural control, deconstructing the nonlinear stability problem into algebraically solvable ones with linear control techniques. This lowers the barrier to extend control-theoretical results to RL. The minimally invasive way allows arguably easy integration into hierarchical RL, and improves its performance by providing consistent low-level dynamic behaviors. This framework also opens opportunities to further integrate with machine learning, in particular, integration with learned latent signals.

Other contributions include (i) illustrating how to transfer the stability of latent signals to physical dynamics with the example of composite variables, and (ii) leveraging task space control to create the latent-space dynamic model composed of the controller gains.

\subsection{Related work.} 

\subsubsection{Contraction theory and safe RL}

Compared with Lyapunov theory, contraction theory analyzes differential dynamics without specifying the system equilibrium and hence robust to the perturbations that shift the system equilibrium \cite{aylward2008stability}, which are common in the real world, particularly for contact-rich manipulation learning. 

Contraction theory has been introduced into learning for robust control \cite{tsukamoto2020neural}, adaptive control \cite{tsukamoto2021learning_adaptive}, motion planning \cite{tsukamoto2021learning}, and system identification \cite{singh2021learning}. 

In safe RL, Krasovskii-Constrained Reinforcement Learning \cite{lale2022kcrl} based on contraction analysis solves the stability constraints with a primal dual policy gradient approach. Another work \cite{sun2020learning} learns a control contraction metric and a tracking controller by optimizing an empirical risk function. 

Our novelty lies in leveraging the modularity of contraction to deconstruct the problem into combinations of algebraically solvable ones. To our best knowledge, this is the first time that the modularity of contraction is introduced into RL. 


Our theoretical results on the constraints for incremental exponential stability are in line with \cite{jin2020stability} based on integral quadratic constraints for $\mathcal{L}_2$ stability, which limits the input gradients of control networks (network Jacobians) for stable linear systems with added nonlinearity. Here we derive the constraints for general nonlinear systems by creating the modularity to deconstruct the nonlinear problem.

Other related work in safe RL, robustness, and generalization can be found in \cite{song2022stability}, the complementary paper that illustrates the necessity of stability constraints for robustness and generalization.

\subsubsection{Modularity} In control, planning, and learning, several threads leverage bioinspired modularity to simplify intrinsically difficult or computationally complex problems. 

The well-known forms of modularity are motor primitives\cite{thoroughman2000learning} and synergies\cite{santello2016hand}. Our framework can be possibly generalized to involve that form of modularity, for example, by expanding the fully actuated 2D formulation of subsystems to underactuated multi-dimensional subsystems. 

Another important direction is modular motion generation, e.g., Riemannian motion policies\cite{ratliff2018riemannian} pairing different Riemannian metrics to the subsets of the state space defined by different mappings from states to the desired acceleration (the polices for motion generation). Our work shares the spirit in the way that we create new Riemannian metrics (made of the coordinate transformation) to define the auxiliary space for dynamic deconstruction. 

The modularity in machine learning\cite{pfeiffer2023modular} is studied for different purposes from the perspective of information theory and computation, formulating the modularity as function compositions and the like. From the perspective of nonlinear system theory, the modularity of contraction theory solves the stability of function compositions with their differential dynamics. We apply those results and analyze stability with differential dynamics, the linear form of which (network Jacobians are embedded in the coefficient matrices) highly simplifies the analysis.

\subsection{Paper organization} 
%
%
%
%
%


 \begin{enumerate}
     \item Dynamic decomposition. 
   	\item Signal composition.
	\item For unknown environments.
	\item Implementation and limitations.
      \item Example 1: necessity for robustness and generalization. 
     \item Example 2: manipulation learning. 
     \item Concluding remarks.
    \item Appendix
 \end{enumerate}

\section{Dynamic decomposition}\label{stability_theorems}

Contraction theory analyzes the differential dynamics to guarantee convergence\cite{lohmiller1998contraction}. Intuitively, provided that the distances between arbitrary trajectories are converging to zero, all possible trajectories are converging to the system limit behaviors, e.g., equilibria or limit cycles.

Provided $\mathbf{\dot y=\mathbf f(\mathbf y, \mathbf u)}$ in the latent space, this section includes (i) contraction from auxiliary space to latent space, (ii) differential dynamics in the auxiliary space, (iii) modular neural control via dynamic decomposition, and (iv) explicit solutions to the hierarchical-type modularity.

\subsection{Contraction from auxiliary space to latent space}

  \begin{figure}[h!]
    \centering
    \includegraphics[width=3.4in]{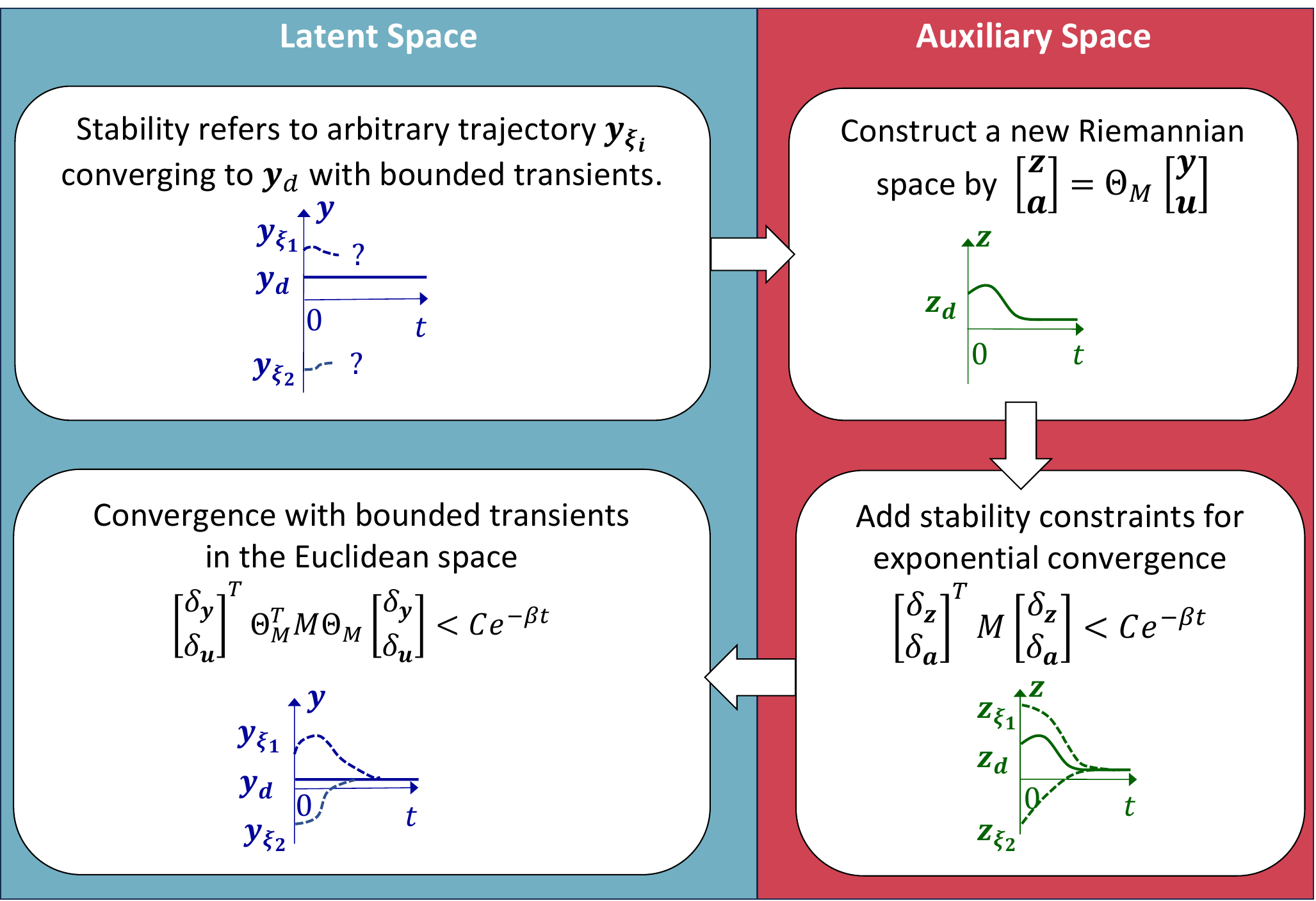}
    \caption{By constraining the dynamics in the auxiliary space to converge exponentially, the dynamics in the Euclidean space is converging with bounded transients.}
    \label{fig:our_methods}
\end{figure}

How the stability is transferred from auxiliary space ($\mathbf z$ and $\mathbf a$) to latent space ($\mathbf y$ and $\mathbf u$) is illustrated in Fig.~\ref{fig:our_methods}. 

Briefly speaking, provided contraction in the auxiliary space, the transformation $T_y$ and $T_a$ maps the contraction metric $M$ in the auxiliary space to the metric $\Theta_M^TM\Theta_M$ in the original latent space, where $\Theta_M$ consists of $T_y$ and $T_a$. Provided that $\Theta_M^TM\Theta_M$ is uniformly positive definite, the dynamics in the original space is also converging with bounded transients. The stability theorem and its proof is in Appendix.~\ref{theorem_1}.

\subsection{Differential dynamics in the auxiliary space}

Here we derive the differential dynamics in the auxiliary space. Note that in the following discussion, we include the integral of $\mathbf z$ in $\mathbf s $ to force the system equilibrium at the goal (Section IV in \cite{song2022stability}). Proofs of dynamic convergence in this paper hold without the integral. 

Given the neural-control system in the original space, by adding two interface layers $T_y$ and $T_a$, we create the new coordinates
\begin{equation}\label{theta_m}
\begin{bmatrix}
\mathbf z \\ \mathbf a 
\end{bmatrix}
= 
\Theta_{M}
\begin{bmatrix}
\mathbf y \\ \mathbf u
\end{bmatrix}
\ \
\text{where} \ \ 
\Theta_M = \begin{bmatrix} T_y & \\  & T_a^{-1}  \end{bmatrix}
\end{equation}
yielding the differential relation 
\begin{equation}\label{zu}
\begin{bmatrix}
\delta \dot{\mathbf{z}} \\ \delta \dot{\mathbf{a}}
\end{bmatrix}
=
\dot\Theta_M 
\begin{bmatrix}
\delta\mathbf y \\ \delta\mathbf u
\end{bmatrix}
+
\Theta_M
\begin{bmatrix}
\delta \dot{\mathbf{y}} \\ \delta \dot{\mathbf{u}}
\end{bmatrix}
\end{equation}
where $\delta \mathbf{y}$, $\delta \mathbf{u}$, $\delta \mathbf{z}$, and $\delta \mathbf{a}$ are differential displacements, i.e., infinitesimal displacements at fixed time.
The control policy $\bm\pi$ can be written as 
\begin{equation}
\mathbf a = \bm\pi(\begin{bmatrix} \mathbf s_1 \\ \mathbf s_2 \end{bmatrix}) =\bm\pi(\begin{bmatrix} \mathbf z \\ \int^t_0 \mathbf zdt\end{bmatrix})
\end{equation}
which yields 
\begin{equation}\label{v}
\delta \dot{\mathbf{a}} 
= 
\frac{\partial \bm\pi}{\partial \mathbf{s}_1}
\delta \dot{\mathbf{z}} +  \frac{\partial \bm\pi}{{\partial \mathbf{s}_2}}
\delta \mathbf{z}
\end{equation}
given $\delta \dot{\mathbf{s}}_2= \delta (\frac{d}{dt}\int^t_0 \mathbf z dt) =  \delta \mathbf z $ and the small change in the input gradient of $\bm\pi$, i.e., $\delta\frac{\partial\bm\pi}{\partial \mathbf s}$, can be ignored.  
The dynamics $\dot{\mathbf y} =\mathbf f(\mathbf y, \mathbf u)$ yields the differential relation 
\begin{equation}\label{e}
\delta \dot{\mathbf{y}} = \frac{\partial \mathbf{f}}{\partial \mathbf{y}}\delta \mathbf{y} + \frac{\partial \mathbf{f}}{\partial \mathbf{u}}\delta \mathbf{u}
\end{equation}

Combining  Eq.~\eqref{zu}, Eq.~\eqref{v}, and Eq.~\eqref{e}, we obtain the differential dynamics 
in the auxiliary space,
\begin{equation}\label{deconstruct_zv}
\begin{bmatrix}
\delta \dot{\mathbf{z}} \\
\delta \dot{\mathbf{a}} 
\end{bmatrix}    
=
\begin{bmatrix}
A & B \\
C & D \\
\end{bmatrix}
\begin{bmatrix}
\delta \mathbf{z} \\
\delta \mathbf{a} 
\end{bmatrix} 
\end{equation}
where
\begin{equation}\label{inverse_solution}
\begin{aligned}
A &= [\dot T_y+T_y\frac{\partial \mathbf{f}}{\partial \mathbf{y}}]T_y^{-1} \\
B &= T_y\frac{\partial \mathbf{f}}{\partial \mathbf{u}}T_a  \\
C &= \frac{\partial \bm\pi}{\partial \mathbf{s}_1} (\dot T_y + T_y \frac{\partial \mathbf{f}}{\partial \mathbf{y}} )T_y^{-1}+ \frac{\partial \bm\pi}{{\partial \mathbf{s}_2}} \\
D &=  \frac{\partial \bm\pi}{\partial \mathbf{s}_1} T_y\frac{\partial \mathbf{f}}{\partial \mathbf{u}} T_a
\end{aligned}
\end{equation}
Note that $\mathbf s_1 =\mathbf z $ and $\mathbf s_2=\int \mathbf z dt$.

\subsection{Modular neural control via dynamic decomposition}\label{theorems}

To illustrate the design of modular neural control, we use $\mathbf z \in \mathbf R^2$ for simplicity, which can be generalized to higher dimensions. 
Given $\mathbf z \in \mathbf R^2$, we rearrange Eq.~\eqref{deconstruct_zv} by grouping $z_i$ and $a_i$, that is, 
\begin{equation}\label{deconstruction_explicit}
\begin{bmatrix}
\delta \dot z_1  \\ \delta \dot a_1  \\ \hline \delta \dot z_2 \\ \delta \dot a_2
\end{bmatrix}
=
\begin{bmatrix}
a_{11} & b_{11} & \aug & a_{12} & b_{12} \\
c_{11} & d_{11} & \aug & c_{12} & d_{12} \\
\hline
a_{21} & b_{21} & \aug & a_{22} & b_{22} \\
c_{21} & d_{21} & \aug & c_{22} & d_{22} \\
\end{bmatrix}
\begin{bmatrix}
\delta z_1  \\ \delta a_1  \\ \hline \delta z_2 \\ \delta a_2
\end{bmatrix}
\end{equation}
where $a_{ij}$, $b_{ij}$, $c_{ij}$ and $d_{ij}$ denote the elements of $A$, $B$, $C$, and $D$ respectively.

\begin{figure}[h!]
            \centering
            \subfigure[Combination of two subsystems.]{\includegraphics[width=3.1in]{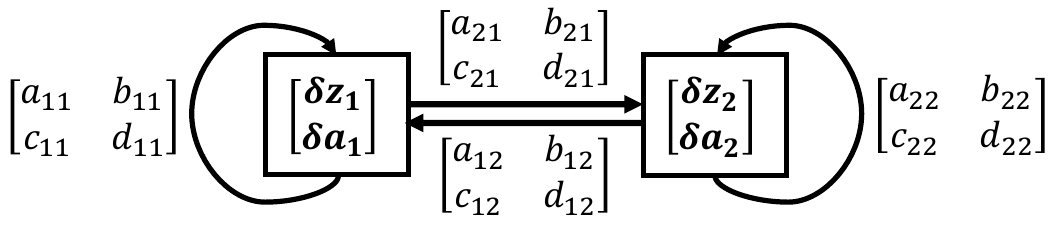}\label{fig:subsystem}}
    \subfigure[Hierarchical combination]{\includegraphics[width=1.3in]{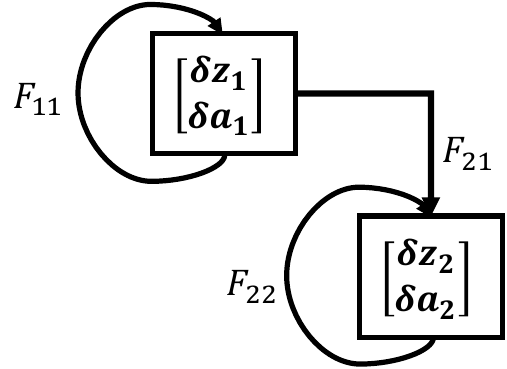}\label{fig:subsys_hierarchical}}  \  \  \
        \subfigure[Feedback combination]{{\includegraphics[width=1.7in]{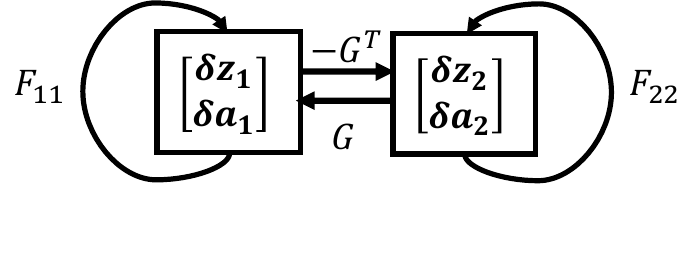}}\label{fig:subsys_feedback}} 
         \caption{Combination examples.}\label{fig:combination_example}
    \end{figure}

The system is then deconstructed into the combinations of two subsystems:
\begin{equation}\label{two_sub}
\begin{aligned}
&
\begin{bmatrix}
\delta \dot z_1  \\ \delta \dot a_1  
\end{bmatrix}
=
\begin{bmatrix}
a_{11} & b_{11} \\
c_{11} & d_{11} \\
\end{bmatrix}
\begin{bmatrix}
\delta z_1  \\ \delta a_1  \end{bmatrix}
+
\begin{bmatrix}
 a_{12} & b_{12} \\
c_{12} & d_{12} \\
\end{bmatrix}
\begin{bmatrix}
\delta z_2 \\ \delta a_2
\end{bmatrix}
\\[8pt]
&
\begin{bmatrix}
 \delta \dot z_2 \\ \delta \dot a_2
\end{bmatrix}
=
\begin{bmatrix}
 a_{22} & b_{22} \\
c_{22} & d_{22} \\
\end{bmatrix}
\begin{bmatrix}
\delta z_2 \\ \delta a_2
\end{bmatrix}
+ 
\begin{bmatrix}
a_{21} & b_{21} \\
c_{21} & d_{21} \\
\end{bmatrix}
\begin{bmatrix}
\delta z_1  \\ \delta a_1  \end{bmatrix}
\end{aligned}
\end{equation}
which is illustrated in Fig.~\ref{fig:subsystem}. The two subsystems are coupled with weight matrices
\begin{equation}\label{coupling}
\begin{bmatrix}
 a_{12} & b_{12} \\
c_{12} & d_{12} \\
\end{bmatrix}
\ \text{and} \
\begin{bmatrix}
 a_{21} & b_{21} \\
c_{21} & d_{21} \\
\end{bmatrix}
\end{equation}
and the self-feedback loops are with weight matrices
\begin{equation}\label{selffeedback}
  \begin{bmatrix}
 a_{11} & b_{11} \\
c_{11} & d_{11} \\
\end{bmatrix}
\ \text{and} \
\begin{bmatrix}
 a_{22} & b_{22} \\
c_{22} & d_{22} \\
\end{bmatrix}
\end{equation}

Given a specific type of combinations, i.e., specified coupling matrices Eq.~\eqref{coupling}, we can solve for $T_y$ and $T_a$ from Eq.~\eqref{inverse_solution} implicitly. In Reference \cite{slotine2003modular}, basic types of combinations that can preserve stability are provided and proved. Here we present two examples: hierarchical and feedback. 

\underline{Hierarchical combination:}
Comparing with the hierarchical form in Section 3.2 of \cite{slotine2003modular} or the illustration in Fig.~\ref{fig:subsys_hierarchical}, it requires $T_y$ and $T_a$ to cancel the upper right block:
\begin{equation}
\begin{bmatrix}
 a_{12} & b_{12} \\
c_{12} & d_{12} \\
\end{bmatrix}
=0
\end{equation}

\underline{Feedback combination:}
Comparing with the feedback form in Section 3.4 of \cite{slotine2003modular} or the illustration in Fig.~\ref{fig:subsys_feedback}, it requires $T_y$ and $T_a$ to make 
\begin{equation}
\begin{bmatrix}
 a_{12} & b_{12} \\
c_{12} & d_{12} \\
\end{bmatrix}
=- 
\begin{bmatrix}
a_{21} & b_{21}\\
c_{21} & d_{21}  \\
\end{bmatrix}^T
\end{equation}

The existence of $T_y$ and $T_a$ needs further study. Intuitively, take the hierarchical-type modularity as an example. There are $2n^2$ unknown variables, that is, the scalar elements from $T_y$ and $T_a$, to satisfy $2(n^2-n)$ equations from specifying the coupling matrices, which implies there always exists a solution. Note that the above discussions are with the assumption that $\dot T_y$ can be ignored (see Appendix.~\ref{F2_assumption}).

Provided that the modularity in control has been realized, the nonlinear stability problem is then simplified to the stability of the self-feedback parts of those subsystems, i.e., the self-feedback matrices Eq.~\eqref{selffeedback}. Solving their characteristic equations for negative eigenvalues yields linear constraints on network Jacobians for stability. Further simplification of the constraints via existing controllers and network architectures are in Section~\ref{imple}, the implementation.

It is possible to solve the coordinate transformation algorithmically. In the following, we provide an explicit solution to the hierarchical-type modularity.

\begin{figure*}[hb!]
    \centering 
        \includegraphics[width=6in]{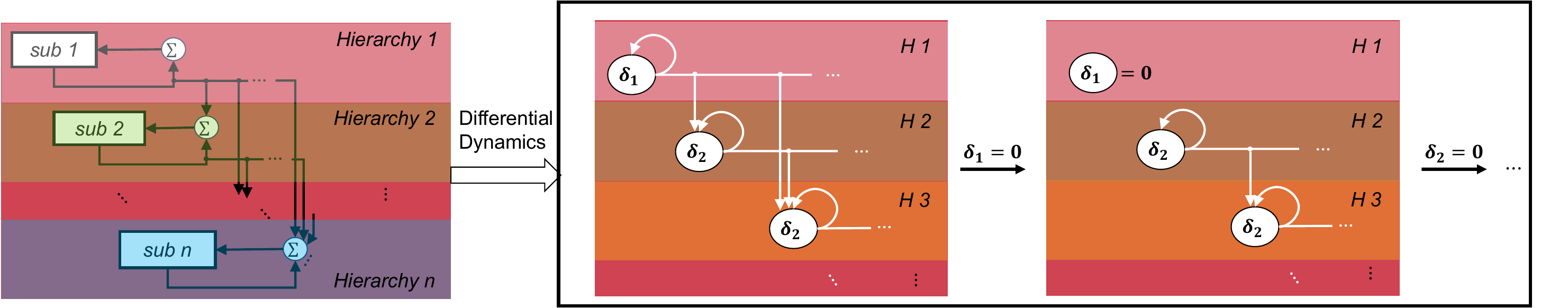}
    \caption{Convergence of hierarchical combinations, where $\bm\delta_i=[\delta z_i, \delta a_i]^T$ denotes the differential displacements between two arbitrary trajectories at fixed time $t$ of the subsystem. When $\bm\delta_i=0$, all possible trajectories of the $i$th subsystem converge to its equilibrium. The subsystems converge recursively following their hierarchies, given stable self-feedback-loops and bounded couplings, as the lowest level with nonzero $\bm\delta$'s is always independent.} 
    \label{fig:combination}
\end{figure*}

\subsection{Explicit solution to the hierarchical-type modularity}\label{stab_theorem_hierarchical}

As for the hierarchical-type modularity, the subsystems converge in the order of their hierarchies, provided that each self-feedback loop is stable and the couplings between subsystems are bounded (Fig.~\ref{fig:combination}). 

\begin{figure}[ht!]
    \centering 
        \includegraphics[width=2.9in]{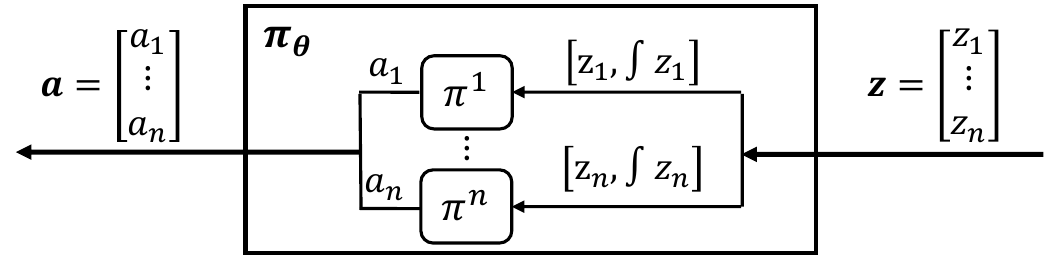}
    \caption{Independent network $\pi^i$ for each dimension $i$.} 
    \label{fig:policy}
\end{figure}

{\underline{Stability Theorem (proof in Appendix~\ref{theorem_2})}:} For the system $\mathbf{\dot y}=\mathbf f(\mathbf y, \mathbf u)$ with diagonalizable $\partial \mathbf f/\partial\mathbf y$, given a neural control policy $\bm \pi$, and two transformation layers $T_y$ and $T_a$  that (i) the control commands $\mathbf u= T_a\bm\pi(T_y\mathbf y)$, and (ii) $T_y^TT_y$ and $(T_a^{-1})^TT_a^{-1}$ are uniformly positive definite, if there exists $\alpha>0$ in a region such that $\forall \mathbf x$ and $\forall t>0$, 
\begin{equation}\label{eq_theorem2}
\begin{aligned}
  \frac{\partial \pi^i}{\partial  s_{1i}}R_{ii} + \Lambda_{ii} <-\alpha,
\ \ \ 
  \frac{\partial \pi^i}{\partial s_{2i}} R_{ii}<-\alpha
\end{aligned}
\end{equation}
where $R_{ii}$ and $\Lambda_{ii}$ are the $i$th diagonal components of $R=T_y\frac{\partial \mathbf{f}}{\partial \mathbf{u}}T_a$ and $\Lambda=T_y\frac{\partial \mathbf{f}}{\partial \mathbf{y}}T_y^{-1}$ respectively, 
the neural control system is contracting in the region, provided that 
\begin{itemize}
\item $T_y$ is the eigenvector matrix from eigenvalue decomposition of diagonalizable $\partial \mathbf{f}/\partial \mathbf{y}$.
\item $T_a=(PQ^T)^{-1}$, where $Q$ is the left matrix from the QR decomposition of $PT_y \partial \mathbf{f}/\partial \mathbf{u}$ and $P$ is the permutation matrix with all ones at the skew diagonal.
\item each subsystem is controlled by an independent neural network, $\bm\pi=[\pi^1, \cdots, \pi^n]$, as illustrated in Fig.~\ref{fig:policy}.
\item there exists $N\in\mathbb R^+$, $\forall t \geq N$, $\dot T_y=0$, and $\forall t <N$, the dynamics are bounded.
\end{itemize}

\section{Signal composition}\label{sec3}

%
%
%
%



The convergence of latent signals does not necessarily imply the stability of physical dynamics. Via composition of latent signals $\mathbf y =\mathbf g(\mathbf x,  \mathbf{\dot x})$, one can enforce a hierarchy of contracting systems again as follows:
\begin{equation}\label{y_and_g}
\begin{aligned}
&\mathbf{\dot y}-\mathbf f(\mathbf y, \mathbf u)=0 \\
&\mathbf g(\mathbf x, \mathbf{\dot x}) -\mathbf y=0
\end{aligned}
\end{equation}
where $\mathbf f$ denotes the resulting dynamics after creating the latent signal $\mathbf y=\mathbf g(\mathbf x, \mathbf{\dot x})$ for the task. 

Provided that the latent signals are converging and the composition of latent signals are stable, i.e., $\mathbf{\dot y}-\mathbf f(\mathbf y, \mathbf u)=0$ and $\mathbf g(\mathbf x, \mathbf{\dot x}) =0$ are contracting, their combined system Eq.~\eqref{y_and_g} is contracting, that is, the stability of the physical dynamics in the task space. 

One example of such stable function composition is the composite variables that mix the tracking errors in position and velocity, $\mathbf y= K_1\mathbf e(t) + K_2\dot{\mathbf e}(t)$ where $\mathbf e=\mathbf x_d-\mathbf x$ and $K_1$ and $K_2$ are positive diagonal matrices. The solution to $\mathbf g(\mathbf x, \mathbf{\dot x}) = K_1\mathbf e(t) + K_2\dot{\mathbf e}(t)=0$ is $\mathbf e=Cexp(-K_1K_2^{-1}t)$ where $C$ is a constant determined by the initial condition. This solution converges exponentially. 

It is possible to integrate with the latent signals in machine learning by $\mathbf y = \bm\pi_{latent}(\mathbf x, \mathbf{\dot x})$ with observable $\mathbf x$ and $\mathbf{\dot x}$, which is not necessarily limited to the physical states of end effectors in the task space. The stability constraints on the input gradients of $\bm\pi_{latent}$ can be derived by applying Theorem 2.1 in \cite{tsukamoto2021contraction}.

%
%
%

\section{For unknown environments}

Our method requires the dynamic model in the latent space, in particular, the partial derivatives $\partial \mathbf f/\partial \mathbf y$ and $\partial \mathbf f/\partial \mathbf u$ in Eq.~\eqref{inverse_solution}. For unknown environments, one may apply system identification to learn the dynamics, $\dot{\mathbf y} = \mathbf f(\mathbf y, \mathbf u)$. Here we present another solution that leverages task space controllers to create a model composed of the controller gains for the applications that can be handled with information from the end effectors. 

Task space controllers in \cite{nakanishi2008operational} can regulate the behaviors at the end effectors to follow some reference $\mathbf p$, for example, acceleration-based $\mathbf p=\mathbf{\ddot x}_r=\mathbf{\ddot x}_d+ K_d (\mathbf{\dot x}_d-\mathbf{\dot x}) + K_p(\mathbf x_d-\mathbf x)$. With the task space controller $\bm\tau=G(\mathbf q, \mathbf {\dot q}, \mathbf p)$, the dynamics of end effectors can be represented as $\mathbf{\ddot e} + K_d\mathbf{\dot e} +K_p\mathbf{e}=0$, where $\mathbf e=\mathbf x_d-\mathbf x$. 

Here we add the RL control into the reference, for example, acceleration-based $\mathbf p =\mathbf{\ddot x}_r + \mathbf u$, leading to the dynamic model in the task space $\mathbf{\ddot e} + K_d\mathbf{\dot e} +K_p\mathbf{e}+\mathbf u=0$. 

Applying the composite variable $\mathbf y= K_1\mathbf e+ K_2\mathbf{\dot e}$, one may create the latent-space model following the equations in Appendix~\ref{feedback_linearization}, leading to the stability constraints as simple as having positive input gradients of networks.

\section{Implementation and limitations}\label{il}

\subsection{Implementation}\label{imple}

The pseudo-code of RL algorithms remains the same, while the graph for control needs to be adapted as follows:

\begin{enumerate}

\item Use the latent signals $\mathbf y=\mathbf g(\mathbf x, \mathbf{\dot x})$ as the RL state for control. 

\item Add two layers $T_y$ and $T_a$ in the control framework as in Fig.~\eqref{fig:model}, the values of which are updated time-stepwise according to the stability theorem in Section.~\ref{stab_theorem_hierarchical}

\item Feed the accumulative error and the current tracking error to the control networks in Fig.~\ref{fig:policy} that each dimension is controlled by an independent network.

\item Limit the network Jacobians $\partial \pi^i/\partial \mathbf s_i$ ($i=1, 2, \cdots, m$) by the linear inequality constraints in Eq.~\eqref{eq_theorem2} where $R_{ii}$ and $\Lambda_{ii}$ updated stepwise. 
\end{enumerate}

\subsection{Further simplification}

When the dynamic without RL control is stable or has been stabilized by existing controllers, i.e., $\sup(\Lambda_{ii})<0$, the linear constraints Eq.~\eqref{eq_theorem2} is further simplified into constraints on the signs of the network Jacobians ($j=1,2$),
\begin{equation}\label{constraint_sign}
\frac{\partial \pi^i}{\partial s_{ji}}R_{ii}<0
\end{equation} 

Further simplification to the signs of network weights can be achieved via network architectures. 

Given MLPs with the hyperbolic tangent function as the activation function, constraints on the signs of network Jacobians are further simplified to constraints on the signs of MLP weights. 

 Assuming an $l$-layered MLP, for each layer, $x_{j+1}=\tanh(W_jx_j+b_j)$ where $j$ denotes the $j^{th}$ layer, $W_j$ is the weights, and $b_j$ is the bias. The partial derivatives $\partial \pi^i/\partial s_{1i}$ and $\partial \pi^i/\partial s_{2i}$ are the products of weights and $[1-\tanh^2{(\cdot)}]$. That is, let $W_{1,1}$ denote the weights of the first layer corresponding to $s_{i1}$, the network Jacobian with respect to $s_{i1}$ becomes
 \begin{equation}\label{product}
\begin{aligned}
\frac{\partial \pi^i}{\partial s_{i1}} = W_lM_{l-1}W_{l-1}\cdots M_2W_2M_1W_{1,1}
\end{aligned}
\end{equation}
where $M_j=1-\tanh^2(W_jx_j+b_j)$. Similarly, one can compute $\partial \pi^i/\partial s_{2i}$. When the activation function is not saturated $M_j\neq 0$, components of $M_j$ are positive. Hence, one can bound the signs of $W_j$ to bound the signs of $\partial \pi^i/\partial s_{ki}$ ($k=1,2$) according to Eq.~\eqref{product} to satisfy the constraints in Eq.~\eqref{constraint_sign}.

\subsection{Limitations}


\paragraph{Limited control frequency} When applying the hierarchical type of modularity in Section.~\ref{stab_theorem_hierarchical}, the control frequency is limited by the eigenvalue decomposition and QR decomposition in solving for $T_y$ and $T_a$. One may design the $\mathbf y=\mathbf g(\mathbf x)$ that makes diagonal $\partial \mathbf f/\partial \mathbf y$ for fast dynamics. 

\paragraph{Loss of optimality} Optimality in control may be affected by forcing the coupling way in the auxiliary space by forcing the function compositions of the latent signals. If applying separate policies to control and planning, the loss of optimality in control can be compensated in planning.

 \section{Example 1: \\ Necessity for robustness and generlization}\label{example1}
 \begin{figure}[h!]
    \centering 
    \includegraphics[width=3.3in]{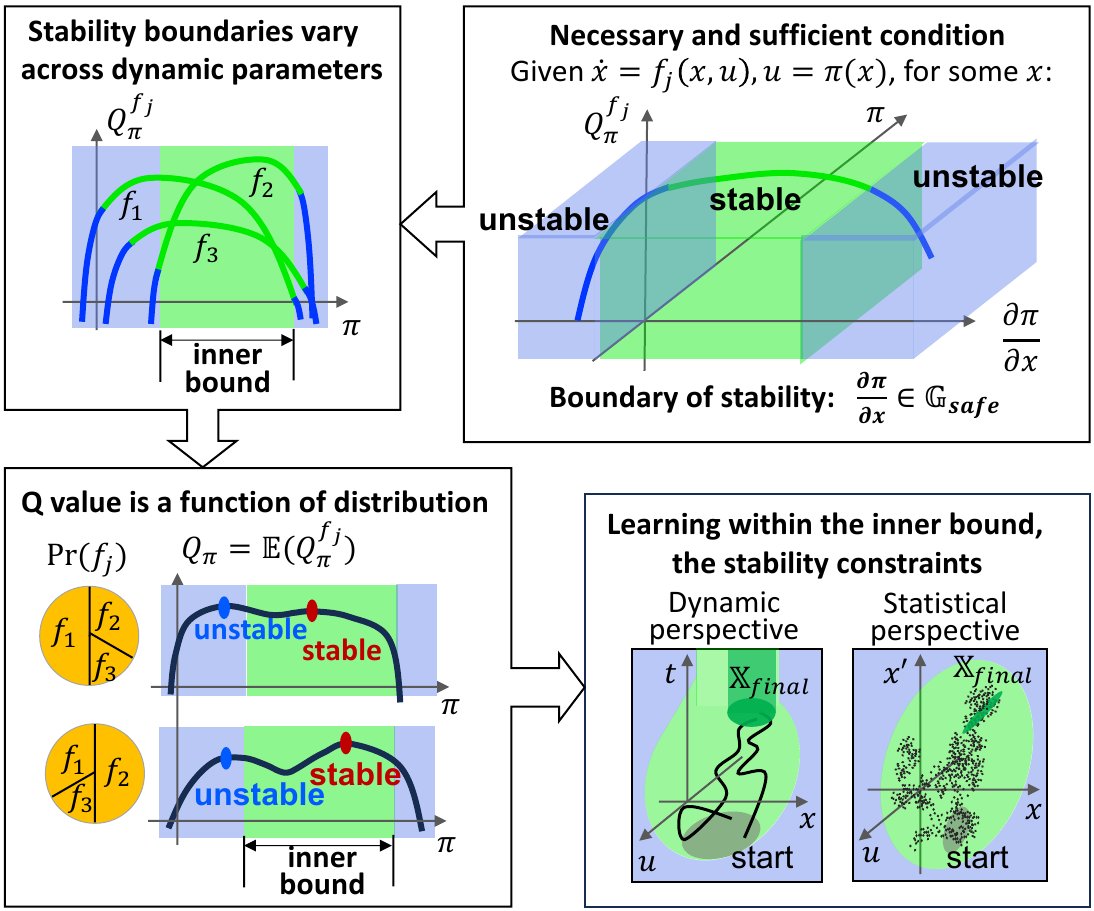}\label{fig:intuition}
    \caption{Intuition. Local optimal policies can contain unstable control because the stability boundaries may not align with the landscape of expected returns.}
\end{figure}

This section briefly presents the main results in \cite{song2022stability}, which illustrates that local optimal policies may contain unstable control for some dynamic parameters and overfitting to such instabilities can deteriorate the generalization of RL. Adding stability constraints can mitigate such concerns.

Intuitively, because the stability boundaries are functions of dynamic parameters and the expected returns are functions of parameter distributions, the boundaries do not necessarily align with the reward (expected returns) landscape, which allows local optimal policies contain unstable control (Fig.~\ref{fig:intuition}).

The example is to move a 2D ``peg" to touch a surface at the desired position $x_d$ with the desired force $f_d$ by proximal policy optimization (PPO)\cite{schulman2017proximal}. The unstable dynamics observed is oscillation and drift, the fraction of which is increasing as learning progresses.  Adding stability constraints via the method here (denoted by C-PPO) can prevent such instabilities and improve generalization (Fig.~\ref{fig:1}). 

The implementation of stable modular control is in Appendix~\ref{PPO}. The latent signals are the state $x$ and force $f$ and a nominal model of the dynamics is provided.

\section{Example 2: Manipulation learning}\label{example2}

To evaluate our method, we created peg-maze and peg-push tasks following the concepts of ant maze and ant push. A Franka Panda robot is used to move the peg. The experiments are in simulation.  

With the easier peg-maze task, we compare our method with the state of the art, a hierarchical learning algorithm HIRO. HIRO plans and tracks in general coordinates with added relabeling processes to alleviate the non-stationary issue, outperforming other methods on ant push and ant maze, particularly in data efficiency\cite{nachum2018data}. 

In our method, we extracted and modified the hierarchical learning framework from the HIRO implementation in Tensorflow Model Garden. In particular, we extracted the 2-level learning framework that includes two TD3 agents. The high-level policy runs with a slower frequency to plan the desired state increments; the low level agent runs faster to track the high-level action within a fixed horizon. 
We removed the relabeling process in HIRO, used the task space instead of the joint space for planning and tracking, and added the composite errors and operational space controllers for unknown environments. 

The resulting framework is illustrated in Fig.~\ref{HRL_model}. The stable neural control framework in Fig.~\ref{fig:model} is implemented in the low level. Note that the integral operator is applied to translation but not orientation errors, as the tasks require strong control for accurate position but prefer flexible orientation. Our method is denoted by TD3/C-TD3. 

The ablation study is performed with the more difficult peg-push task, to evaluate the effect from neural controllers and from stability constraints. We compare our TD3/C-TD3 with 
\begin{itemize}
\item One-level learning with non-neural control: 

TD3/Control and PPO/Control, 
\item Hierarchical learning without stability constraints: 

TD3/TD3. 
 \end{itemize}

 \subsection{TD3/C-TD3 vs. HIRO on peg maze.}

\paragraph{Task setting.} The task is to reach a goal in the green zone through the unstructured environment with fixed obstacles (top in Fig.~\ref{HRL_env}). Robot's motion is limited by a hood. Particularly, the vertical limit forces the peg going through the unstructured zone. 

\paragraph{Experiment}
Both HIRO and TD3/C-TD3 can find the path with high success rates and with high data efficiency ($\sim$ 2M and 1M respectively) given the initial exploration in the direction of the maze. To evaluate the safety in exploration, we tested policies for TD3/C-TD3 at 10k, 200k, 400k, 600k, 800k, and 1M steps (training converged), and for HIRO at 10k, 500k, 1M, 1.5M, 2M, and 2.3M steps (training converged); and we compare the distribution of contact forces.  

The distribution of contact forces is plotted in Fig.~\ref{peg_maze_results}. In TD3/C-TD3, $99.93\%$ steps are with less-than-1 N contacts, including $64.24\%$ steps without contacts. In HIRO,  $65.88\%$ steps are with less-than-1 N contacts, including $6.22\%$ zero contacts.  Trajectory examples are in \url{https://www.youtube.com/watch?v=nFLHwVfPIJw}. This shows that TD3/C-TD3 can explore through the obstacles with soft touches, i.e., small contact forces. 

Besides the stable task-space neural control in the low level, TD3/C-TD3 also benefits from an easier high-level planning problem: finding a path vs. planning joint positions in HIRO.

\subsection{Ablation study: peg-push task}

\begin{figure*}[ht!]
            \centering
             \  \ \ \ \ \ \ \  \  \
            \subfigure[Hierarchical RL with stable modular neural control in the low level (TD3/C-TD3).]{{\includegraphics[width=4.1in]{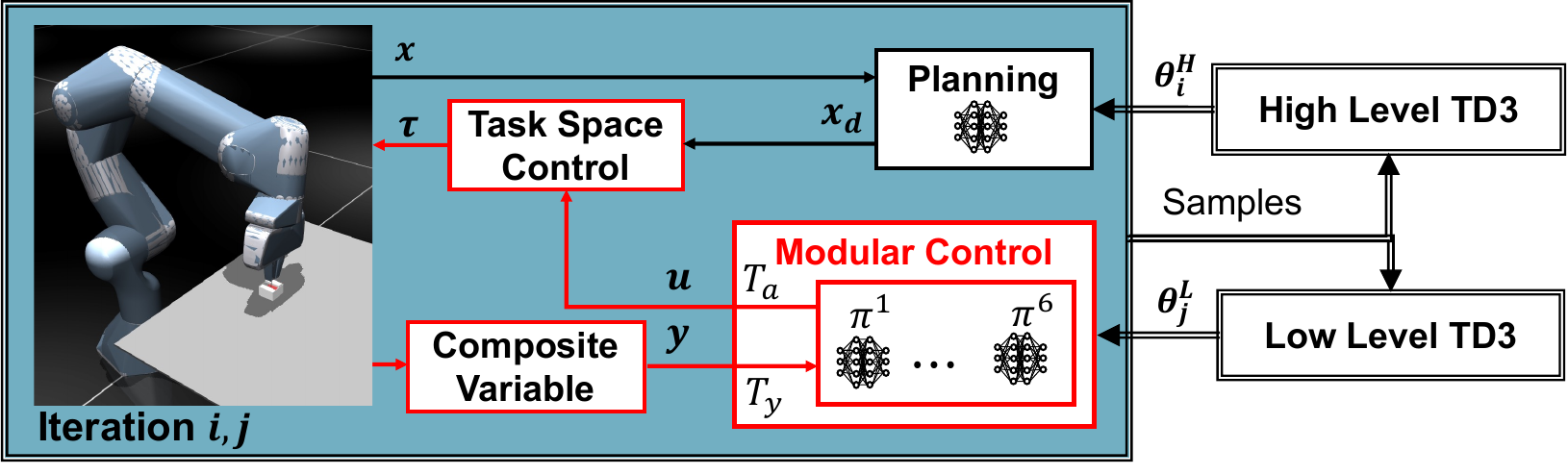}}\label{HRL_model}} \  \ \  \ \ \ \ 
    \subfigure[Peg maze and peg push.]{{\includegraphics[width=1.9in]{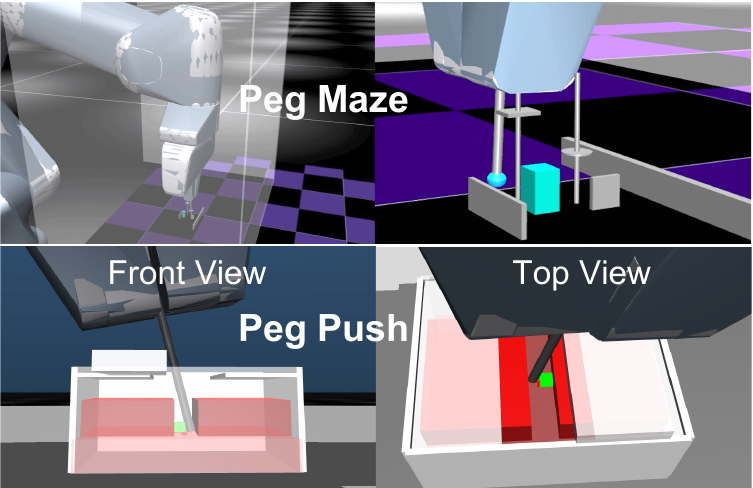}}\label{HRL_env}} \\
    \subfigure[Peg maze: Contact force distributions and the success rates.]{{\includegraphics[width=5.1in]{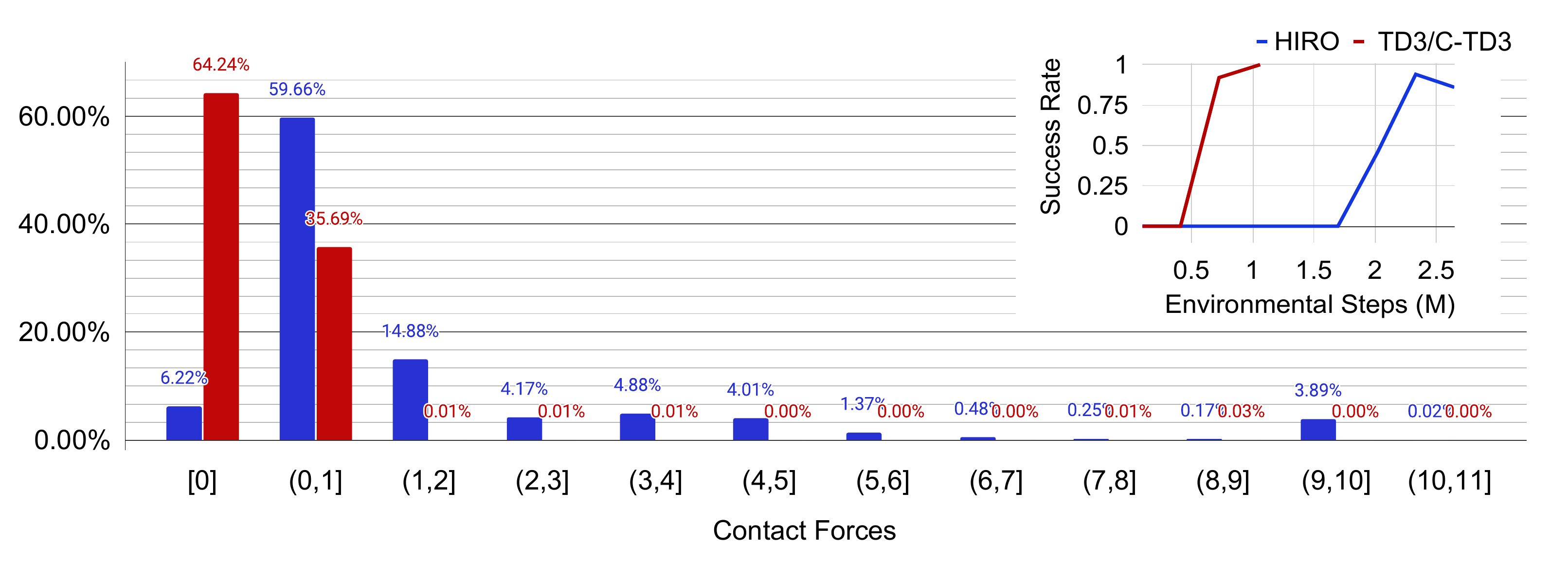}}\label{peg_maze_results}} \ \ 
        \subfigure[{Peg push: Learning curves.}]{{\includegraphics[width=1.9in]{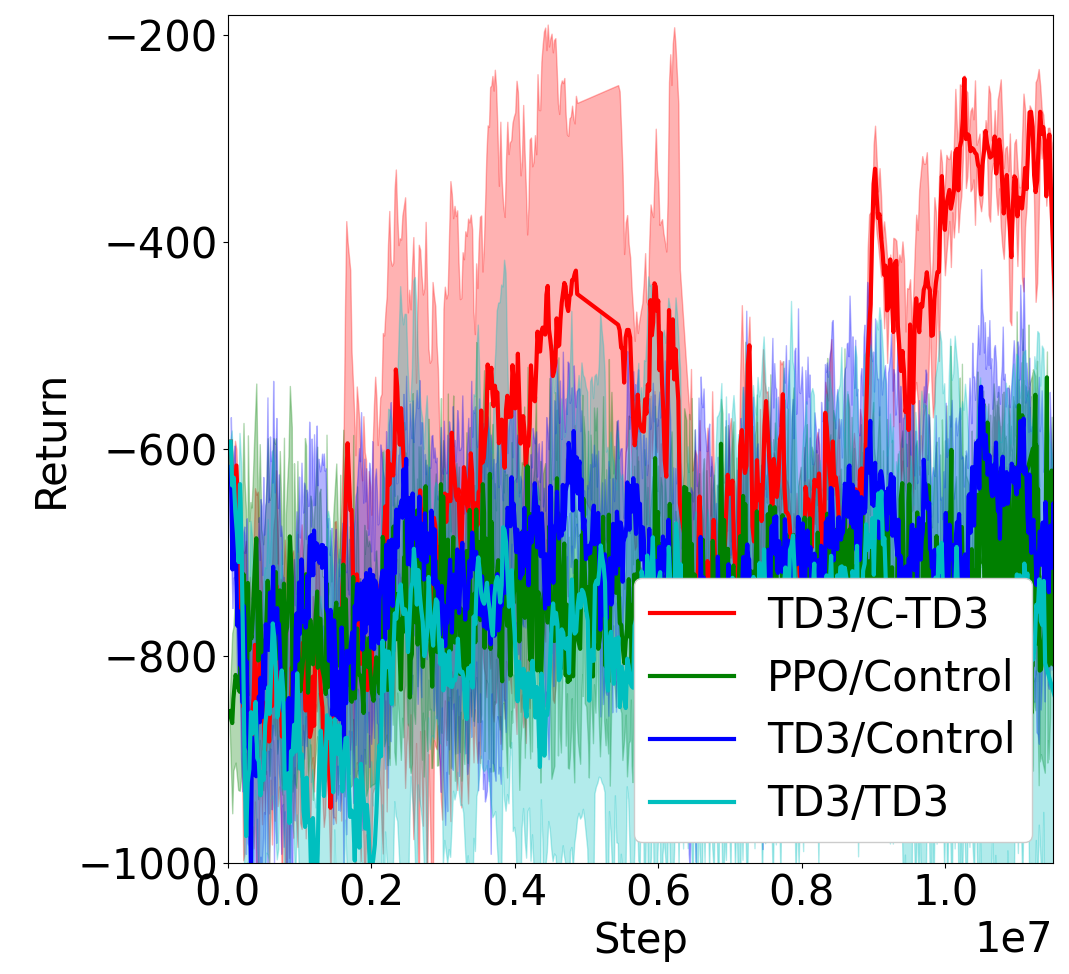}}\label{peg_push_results}} 
         \caption{Effectiveness in manipulation learning (trajectory examples in \url{https://www.youtube.com/watch?v=nFLHwVfPIJw} for peg maze and \url{https://www.youtube.com/watch?v=RpGzpZPifUw} for peg push.) (a) Hierarchical RL framework of TD3/C-TD3. (b) Peg maze and peg push tasks. Peg maze (top) is a goal reaching task with fixed obstacles both near the goal (randomly sampled in the green zone) and near the robot arm (the transparent hood). Peg push (bottom) is a goal reaching task that requires opening a sliding lid box, inserting into a deep slot, and pushing away the green obstacle to reach the small red spot underneath. (c) Peg maze results: TD3/C-TD3 vs. HIRO (the state-of-the-art data-efficient hierarchical RL). Both TD3/C-TD3 and HIRO can quickly learn the task ($\sim$1M and 2M steps). The contact force distribution are estimated from policies (success rates on top right) at 10k, 200k, 400k, 600k, 800k, and 1M steps for TD3/C-TD3, and at 10k, 500k, 1M, 1.5M, 2M, and 2.3M steps for HIRO. The distribution shows improvements in safe exploration that in TD3/C-TD3, $99.93\%$ are with less than 1 N, and $65.88\%$ with HIRO. (d) Peg push results: TD3/C-TD3 vs. RL with non-neural controllers (PPO/Control, TD3/Control) vs. TD3/TD3 without stability constraints. Only TD3/C-TD3 successfully learn the task within 12 M steps, with 2 trials learned the task around 3 M steps causing the large variation. TD3/control and PPO/control can learn stable trajectories but haven’t figured out the entire path to the goal. Without stability constraints, TD3/TD3 learns unstable trajectories passing by the goal.}\label{fig:hiro_exp}
\end{figure*}

\paragraph{Task setting}
The peg-push task is to reach a hidden goal in a box (bottom in Fig.~\ref{HRL_env}) that requires interactions with moving objects. The robot needs to push open the sliding lid, insert into a deep slot, and push away the green obstacle to reach the small red spot. Appendix~\ref{HRL_example} lists the geometry and contact settings.

\paragraph{Experiment}
We firstly tuned the task space controller, the high-level action range, the low-level horizon, and the fixed trajectory length to make sure that the robot can stably explore the entire environment. Controller gains, RL hyperparameters, and other task parameters are listed in Appendix~\ref{HRL_example}. 

The average learning curve from 5 seeds are presented in Fig.~\ref{peg_push_results}. Only TD3/C-TD3 is able to learn the task (expected return in $[-400, -200]$), and 2 trials have converged after 3M experiment steps. This causes the large variance around 3M to 6M in the averaged learning curve. Other methods fail to learn the task (returns $\approx -700$). 

Close examination of the trajectories shows that TD3/control and PPO/control can learn stable trajectories but haven't figured out the entire path to the goal within 12M steps. Without stability constraints, TD3/TD3 learns unstable trajectories passing by the goal. Trajectory examples can be found at \url{https://www.youtube.com/watch?v=RpGzpZPifUw}. 

There still exists one question: whether the soft contact model highly simplifies the task as we observed penetration (the penetration issue discussed in \cite{parmar2021fundamental}). Further study will focus on evaluation of the neural controller for complex contact stiffness that varies across different surface materials and changes along with relative motions. 

\section{Concluding Remarks}
Bioinspired modularity can deconstruct intrinsic difficulties and computational complexities in control, planning, and learning. We realize modular neural control for RL via contraction theory, leading to a minimally invasive way for stability guarantees, which allows arguably easy integration with the modular neural architectures in machine learning, in particular, hierarchical RL, and improves its performance. This modular control framework provides a new perspective to integrate control theoretical results with RL: applying control techniques to signal composition and dynamic decomposition for modularity. Towards full integration with the modular neural architectures in machine learning, the concept applies to dynamics beyond physical robotics.

\section{Appendix}
The appendix at the end of this paper includes the following sections. 
\begin{enumerate}[label=\Alph*.]
\item Incremental exponential stability.
\item Stability theorem for smooth coordinate transformation.
\item Proof of the theorem for the hierarchical combination. . 
\item Integration of composite variables and task space controllers for unknown environments. 
\item Robustness. 
\item Implementation in the PPO example
\item Peg maze example. 
\item Peg push example. 
\end{enumerate}

\begin{figure*}[ht!]
    \centering 
    \subfigure[Peg touching by PPO.]{\includegraphics[width=1.6in]{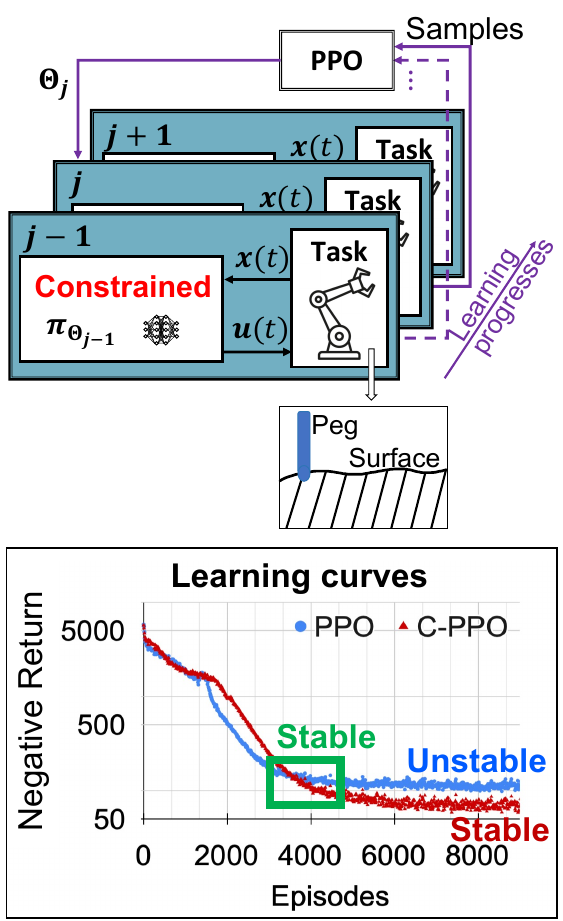}\label{fig:concept}} 
    \subfigure[Stability: mitigation of oscillation and drift by constraints]{\includegraphics[width=3.83in]{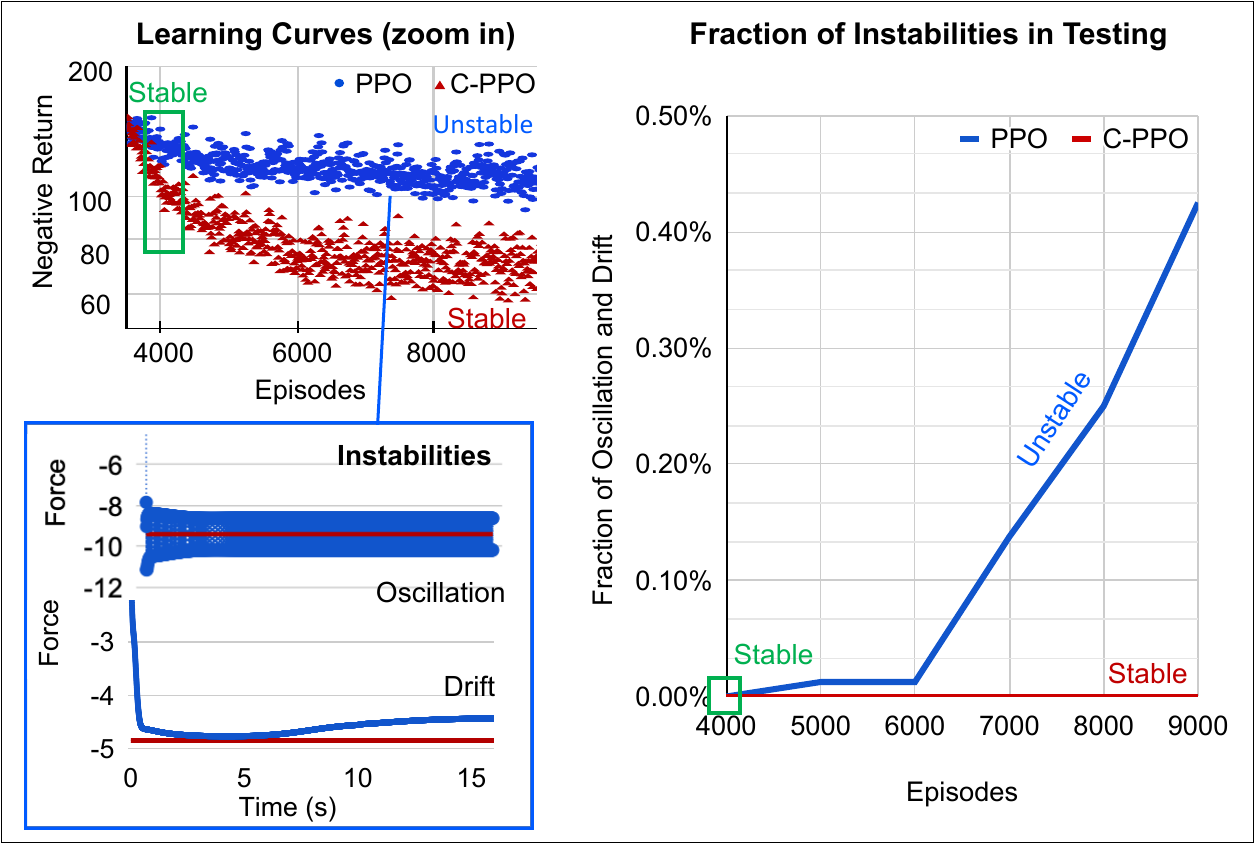}\label{fig:traj}}
    \subfigure[Generalization and robustness.]{{\includegraphics[width=1.59in]{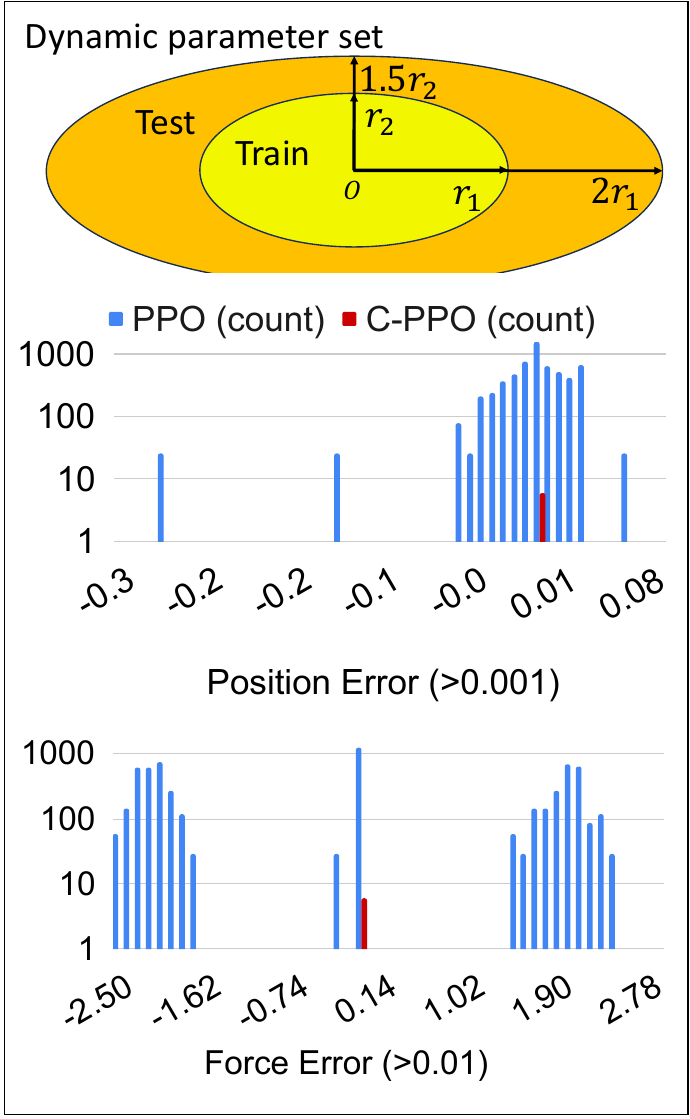}}\label{ppo_stability}} 
    \caption{Necessity for robustness and generalization (more results in \cite{song2022stability}). (a) 2D peg-touching task. Both PPO and C-PPO can learn the task with similar return levels and with similar data efficiency. (b) Stability. Starting from the policy at Episode 5000, 8000 trajectories are tested every 1k episodes. In PPO, the amount of oscillation and drift is increasing as learning progresses, which implies overfitting, while in C-PPO, the fraction remains zero. (c) Generalization and robustness. The policies at Episode 5000 are tested for 8000 trajectories, with new task parameters with $50\%$ differences in $\tau_i$ ($i=x,z$), $100\%$ in $K_{sur}$ and $50\%$ in $K_i$ ($i=1,2$) compared to the training set. The nominal model from the training set is used in C-PPO. The histogram of tracking errors shows dramatic improvements in generalization in C-PPO.}\label{fig:1}
\end{figure*}

\section*{Acknowledgments} This study was supported in part under the RIE2020 Industry Alignment
  Fund – Industry Collaboration Projects (IAF-ICP) Funding Initiative,
  as well as cash and in-kind contribution from the industry partner,
  HP Inc., through the HP-NTU Digital Manufacturing Corporate Lab.

\bibliographystyle{IEEEtran}
\bibliography{refs}




\balance

\newpage

\nobalance

\section{Appendix}

\subsection{Incremental exponential stability}\label{incre_sta}

The stability here in this paper refers to the incremental exponential stability as follows.

\underline{Definition\cite{tsukamoto2021contraction}.}
Let $\xi_1(t)$ and $\xi_2(t)$ denote two trajectories of a system $\dot{\mathbf{x}}=\mathbf{g}(\mathbf{x},t)$. This system $\dot{\mathbf{x}}=\mathbf{g}(\mathbf{x},t)$ is incrementally exponentially stable if there exist $C>0$ and $\beta>0$ such that 
\begin{equation}\label{xi}
    \lVert \xi_1(t)-\xi_2(t) \rVert \leq Ce^{-\beta t}  \lVert \xi_1(0)-\xi_2(0) \rVert 
\end{equation}

Note that stability of nonlinear systems can be defined in multiple ways and the differences are discussed in \cite{tsukamoto2021contraction}.

\subsection{Stability theorem for smooth coordinate transformation}\label{theorem_1}

{\underline{Theorem}:} 
Given a dynamic system $\dot{\mathbf y} =\mathbf f(\mathbf y, \mathbf u, t)$, a neural control policy $\bm \pi$, and two transformation layers $T_y$ and $T_a$  that (i) the control commands $\mathbf u= T_a\bm\pi(T_y\mathbf y)$, and (ii) $T_y^TT_y$ and $(T_a^{-1})^TT_a^{-1}$ are uniformly positive definite, this neural control system is incrementally exponentially stable in a region, if $\forall t>0$, $\forall \mathbf y$ in the region, there exists $\beta>0$ such that
\begin{equation}\label{F1_theorem1}
(F_1^T+F_1)\prec -[\beta + max(\nu^+,0)]I
\end{equation}
where 
\begin{equation}\label{F1}
F_1 = \begin{bmatrix}
T_y\frac{\partial \mathbf{f}}{\partial \mathbf{y}}T_y^{-1} & T_y\frac{\partial \mathbf{f}}{\partial \mathbf{u}} T_a \\[8pt]
\frac{\partial \bm\pi}{\partial \mathbf{s}_1}
T_y\frac{\partial \mathbf{f}}{\partial \mathbf{y}}T_y^{-1} + \frac{\partial \bm\pi}{{\partial \mathbf{s}_2}} & \frac{\partial \bm\pi}{\partial \mathbf{s}_1}
T_y\frac{\partial \mathbf{f}}{\partial \mathbf{u}} T_a 
\end{bmatrix}
\end{equation}
and $\nu^+$ is the upper bound of the eigenvalues of $F_2^T+F_2$, 
\begin{equation}\label{F2}
F_2 = 
\begin{bmatrix}
\dot T_yT_y^{-1} & 0 \\[8pt]
\frac{\partial \bm\pi}{\partial \mathbf{s}_1}\dot T_yT_y^{-1} & 0
\end{bmatrix} 
\end{equation}
and $\mathbf s=[\mathbf s_1^T, \mathbf s_2^T]^T=[\mathbf z^T, \int_0^t\mathbf z^Tdt]^T$ with $\mathbf z=T_y\mathbf y$. 

{\underline{Proof}:} 


\paragraph{Differential dynamics in the auxiliary space} 
Provided that $\mathbf z = T_y\mathbf y$ and $\mathbf a = T_a^{-1}\mathbf u$, the differential dynamics in the auxiliary space is Eq.~\eqref{deconstruct_zv}, which can also be represented as 
\begin{equation}\label{delta_zv}
\begin{bmatrix}
\delta \dot{\mathbf{z}} \\
\delta \dot{\mathbf{a}} 
\end{bmatrix}    
=
F
\begin{bmatrix}
\delta \mathbf{z} \\
\delta \mathbf{a} 
\end{bmatrix} 
\end{equation}
where 
\begin{equation}\label{F}
F= \begin{bmatrix}
[\dot T_y+T_y\frac{\partial \mathbf{f}}{\partial \mathbf{y}}]T_y^{-1} & T_y\frac{\partial \mathbf{f}}{\partial \mathbf{u}}T_a \\[8pt]
 F_{21} &  \frac{\partial \bm\pi}{\partial \mathbf{s}_1} T_y\frac{\partial \mathbf{f}}{\partial \mathbf{u}} T_a
\end{bmatrix}
\end{equation}
and
\begin{equation}
F_{21}=\frac{\partial \bm\pi}{\partial \mathbf{s}_1} (\dot T_y + T_y \frac{\partial \mathbf{f}}{\partial \mathbf{y}} )T_y^{-1}+ \frac{\partial \bm\pi}{{\partial \mathbf{s}_2}}
\end{equation}

\paragraph{Sufficient condition that makes the squared distance between trajectories exponentially converge in the auxiliary space}
The time derivative of the squared distance in the auxiliary space can be written as 
\begin{equation}
\begin{aligned}
&\frac{d}{dt}\Big(\begin{bmatrix} \delta \mathbf z \\ \delta\mathbf  a \end{bmatrix}^T\begin{bmatrix} \delta \mathbf z \\ \delta \mathbf a \end{bmatrix}\Big) 
= 2\begin{bmatrix} \delta \mathbf z \\ \delta \mathbf a \end{bmatrix}^T\frac{d}{dt}\begin{bmatrix} \delta \mathbf z \\ \delta \mathbf a \end{bmatrix} \\[8pt]
=&2\begin{bmatrix} \delta \mathbf z \\ \delta \mathbf a \end{bmatrix}^T  F\begin{bmatrix} \delta \mathbf z \\ \delta \mathbf a \end{bmatrix} 
\leq  2\lambda_{max}(F)\Big(\begin{bmatrix} \delta \mathbf z \\ \delta \mathbf a \end{bmatrix}^T\begin{bmatrix} \delta \mathbf z \\ \delta \mathbf a \end{bmatrix}\Big)
\end{aligned} 
\end{equation}
where $\lambda_{max}(F)$ represents the function that takes the largest eigenvalue of $\frac{1}{2}(F^T+F)$. 

If  $\lambda_{max}(F)$ is uniformly strictly negative, that is, $\exists \beta>0$, $\forall \mathbf x, \forall t>0, \lambda_{max}(F)\leq -\beta<0$, 
\begin{equation}
\begin{aligned}
&(\delta\mathbf z)^T( \delta\mathbf z )+( \delta \mathbf a)^T (\delta\mathbf a)  \\[8pt]
 \leq &e^{\int_0^t2\lambda_{max}(F)dt} [(\delta\mathbf z_0)^T( \delta\mathbf z_0 )+( \delta \mathbf a_0)^T (\delta\mathbf a_0) ]
\\[8pt]
 = &e^{-2\beta t}  [(\delta\mathbf z_0)^T( \delta\mathbf z_0 )+( \delta \mathbf a_0)^T (\delta\mathbf a_0) ]
\end{aligned}
\end{equation}
where $\delta\mathbf z_0$ and $\delta \mathbf a_0$ denotes the initial values at $t=0$

\paragraph{Distance between two arbitrary trajectories}

The uniformly positive definite $T_y^TT_y$ means that $\exists \underline m>0$, $\forall \mathbf y, \forall t>0$, $T_y^TT_y\succeq \underline m I$.  Similarly, the uniformly positive definite $(T_a^{-1})^TT_a^{-1}$ makes $( \delta \mathbf u)^T(T_a^{-1})^TT_a^{-1}(\delta\mathbf u) >0$. 

Given two arbitrary trajectories $\xi_0$ and $\xi_1$, the distance is defined as the smallest path integral between two points at some fixed time $t$, that is, 
\begin{equation}
\begin{aligned}
&\Vert \xi_1(t)-\xi_0(t) \Vert 
=  \int_{\xi_0}^{\xi_1}\Vert \delta \mathbf y\Vert \\[8pt]
= & \int_{\xi_0}^{\xi_1}\frac{\sqrt{(\delta\mathbf y)^T\underline{m} I\delta\mathbf y }}{\sqrt{\underline m}}   \\[8pt]
\leq  &\int_{\xi_0}^{\xi_1}\frac{\sqrt{(\delta\mathbf y)^TT_y^TT_y\delta\mathbf y }}{\sqrt{\underline m}}  \\[8pt]
\leq &\int_{\xi_0}^{\xi_1}\frac{\sqrt{(\delta\mathbf y)^TT_y^TT_y\delta\mathbf y + ( \delta \mathbf u)^T(T_a^{-1})^TT_a^{-1}(\delta\mathbf u)  }}{\sqrt{\underline m}} \\[8pt]
= &\int_{\xi_0}^{\xi_1} \sqrt{\frac{ [(\delta\mathbf z)^T( \delta\mathbf z )+( \delta \mathbf a)^T (\delta\mathbf a) ]}{\underline m}} \\[8pt]
\leq &\int_{\xi_0}^{\xi_1} \sqrt{\frac{e^{-2\beta t}  [(\delta\mathbf z_0)^T( \delta\mathbf z_0 )+( \delta \mathbf a_0)^T (\delta\mathbf a_0) ]}{\underline m}} \\[8pt]
=&Ce^{-\beta t}  \lVert \xi_1(0)-\xi_2(0) \rVert 
\end{aligned}
\end{equation}
where $C=\int_{\xi_0}^{\xi_1} \sqrt{\frac{[(\delta\mathbf z_0)^T( \delta\mathbf z_0 )+( \delta \mathbf a_0)^T (\delta\mathbf a_0)}{\underline m(\delta\mathbf x_0)^T( \delta\mathbf x_0)}} $ that is a constant determined by the initial values at $t=0$.

Therefore, if $\lambda_{max}(F)$ is uniformly strictly negative in some region, the feedback-control system is incrementally exponentially stable. 

\paragraph{Boundary for the constant part in $F$}
We rewrite $F$, Eq.~\eqref{F}, into the sum of the constant $F_1$, Eq.~\eqref{F1}, and the time-varying $F_2$, Eq.~\eqref{F2}. Let $\nu^+$ denote $\lambda_{max}(F_2)$. If there exists $\beta>0$,  $ \lambda_{max}(F_1)<- [\beta+\max(\nu^+,0)]$,
\begin{equation}
\begin{aligned}
\lambda_{max}(F)<&\lambda_{max}(F_1) + \lambda_{max}(F_2) \\
<&-[\beta+\max(\nu^+,0)]+\nu^+ \\
\leq &-\beta
\end{aligned}
\end{equation}

Hence, if $\forall t>0$, $\forall \mathbf y$ in a region, $\exists \beta>0$,  $\lambda_{max}(F_1)<- [\beta+\max(\nu^+,0)]$, the feedback-control system in Fig.~\ref{fig:model} is incrementally exponentially stable in the region. This region is a contraction region.

Proof is finished.

\subsection{Proof of the theorem for the hierarchical combination.}\label{theorem_2}

\subsubsection{Assumption that the term involving $F_2$ in Eq.~\eqref{F2} can be seen as bounded temporary disturbances}\label{F2_assumption}

The assumption that the term involving $F_2$ in Eq.~\eqref{F2} can be seen as bounded temporary disturbances means there exists $N>0$, 
\begin{itemize}
\item $\forall t\geq N$, $\dot T_y=0$
\item $\forall t<N$, there exists upper bound $C_N$ of the distance between two arbitrary trajectories $  \lVert \xi_1(t)-\xi_2(t) \rVert < C_N$
\end{itemize}

\subsubsection{Differential dynamics in the auxiliary space}

With the above assumption, the dynamic stability in the auxiliary space Eq.~\eqref{delta_zv} can be analyzed with the model
\begin{equation}\label{delta_zv_F1}
   \begin{bmatrix}
\delta \dot{\mathbf{z}} \\
\delta \dot{\mathbf{a}} 
\end{bmatrix}    
= F_1\begin{bmatrix}
\delta \mathbf{z} \\
\delta \mathbf{a} 
\end{bmatrix}  
\end{equation}
because given Eq.~\eqref{delta_zv_F1} is contracting with the rate $\beta>0$ in Eq.~\eqref{xi}, Eq.~\eqref{delta_zv} is also contracting with the rate $\beta$:
\begin{itemize}
\item $\forall t\geq N$, Eq.~\eqref{delta_zv} with $\dot T_y=0$ becomes Eq.~\eqref{delta_zv_F1},
\item $\forall t<N$, the distance between two arbitrary trajectories is also bounded, $\lVert \xi_1(t)-\xi_2(t) \rVert < (C_Ne^{\beta N})e^{-\beta t}$, where $C_N$ denotes the upper bound of the distance for $t<N$. 
\end{itemize}

\subsubsection{Diagonal $\Lambda$ via $T_y$}

Given diagonalizable $\partial \mathbf f/\partial \mathbf y$ and $T_y$ being the eigenvector matrix,\begin{equation}
    \Lambda = T_y\frac{\partial \mathbf{ f}}{\partial \mathbf{y}}T_y^{-1} =
    \begin{bmatrix}
        \Lambda_{11} &  &  & \\
         & \Lambda_{22} & &  \\
         & & \ddots & \\
         & & & \Lambda_{nn} 
    \end{bmatrix}
\end{equation}
where $\Lambda_{ii}$ denotes the eigenvalues.

\subsubsection{Upper triangular $R$ via $T_a$}
The term $R$ equals to $PR_{qr}^TP$, which is an upper triangular matrix that can be represented as
  \begin{equation}
    R = T_y\frac{\partial \mathbf{f}}{\partial \mathbf{u}}T_a
    =  \begin{bmatrix}
        R_{11} & R_{12} & \cdots & R_{1n} \\
         & R_{22} & & R_{2n} \\
         & & \ddots & \vdots \\
         & & & R_{nn} 
    \end{bmatrix}  
  \end{equation}  
because 
    \begin{equation}
    \begin{aligned}
    R=&T_y\frac{\partial \mathbf{f}}{\partial \mathbf{u}}T_a\\
    =&(PP)T_y\frac{\partial \mathbf{f}}{\partial \mathbf{u}}(PQ^T)^{-1} \\
    =& P[(T_y\frac{\partial \mathbf{f}}{\partial \mathbf{u}})^TP]^T(PQ^T)^{-1} \\
    =&P(QR_{qr})^T(Q^T)^{-1}P \\
    =& PR_{qr}^TQ^T(Q^T)^{-1}P \\
    =&PR_{qr}^TP \\
    \end{aligned}
    \end{equation} 
since $T_a=(PQ^T)^{-1}$, $Q$ is the left matrix in the QR decomposition $QR_{qr}=(T_y\frac{\partial \mathbf{f}}{\partial \mathbf{u}})^TP$ with    
\begin{equation}
    P = \begin{bmatrix}
     & & 1 \\
    & \iddots &  \\
    1 & & 
    \end{bmatrix}
    \end{equation}

\subsubsection{Hierarchical combination of subsystems}

Plugging $\Lambda=T_y\frac{\partial \mathbf{f}}{\partial \mathbf{x}}T_y^{-1}$ and $R=T_y\frac{\partial \mathbf{f}}{\partial \mathbf{u}}T_a$ into Eq.~\eqref{delta_zv_F1}, the dynamics then can be written as 
\begin{equation}\label{zv}
\begin{aligned}
   \begin{bmatrix}
\delta \dot{\mathbf{z}} \\
\delta \dot{\mathbf{a}} 
\end{bmatrix}    
&=
\begin{bmatrix}
\Lambda & R \\[8pt]
\frac{\partial \bm\pi}{\partial \mathbf{s}_1}\Lambda + \frac{\partial \bm\pi}{{\partial \mathbf{s}_2}} & \frac{\partial \bm\pi}{\partial \mathbf{s}_1}
R
\end{bmatrix}
\begin{bmatrix}
\delta \mathbf{z} \\
\delta \mathbf{a} 
\end{bmatrix}
\end{aligned} 
\end{equation}
where 
\begin{equation}
\begin{aligned}
    \Lambda &= T_y\frac{\partial \mathbf{ f}}{\partial \mathbf{y}}T_y^{-1} =
    \begin{bmatrix}
        \Lambda_{11} &  &  & \\
         & \Lambda_{22} & &  \\
         & & \ddots & \\
         & & & \Lambda_{nn} 
    \end{bmatrix}
    \\[8pt]
    R &= T_y\frac{\partial \mathbf{f}}{\partial \mathbf{u}}T_a
    =  \begin{bmatrix}
        R_{11} & R_{12} & \cdots & R_{1n} \\
         & R_{22} & & R_{2n} \\
         & & \ddots & \vdots \\
         & & & R_{nn} 
    \end{bmatrix}
\end{aligned}
\end{equation}

With the policy that is made of a sequence of networks in parallel in Fig.~\ref{fig:policy}, 
the dynamics Eq.~\eqref{zv} can be written explicitly with $F_1$ being represented by two column blocks ($F_1=[F_{1l} \ F_{1r}]$) as follows:
\begin{equation}
\begin{bmatrix}
\delta \dot z_1 \\ \delta \dot z_2  \\
\vdots \\
\delta \dot z_n \\ \delta \dot a_1  \\
\delta \dot a_2 \\
\vdots \\
\delta \dot a_n  
\end{bmatrix}
=
\begin{bmatrix}
F_{1l} & F_{1r}
\end{bmatrix}
\begin{bmatrix}
\delta  z_1 \\ \delta z_2  \\
\vdots \\
\delta z_n \\ \delta a_1  \\
\delta a_2 \\
\vdots \\
\delta a_n  
\end{bmatrix}
\end{equation}
where 
\begin{equation}
\begin{aligned}
& F_{1l}
=
\begin{bmatrix}
 \Lambda_{11} &   &  \\
  &   \ddots &  \\
   &  &  \Lambda_{nn}  \\[10pt]
  \frac{\partial \pi^1}{\partial s_{11}} \Lambda_{11} + \frac{\partial \pi^1}{\partial s_{21}} &  & \\
 &  \ddots &  \\
   &  &  \frac{\partial \pi^n}{\partial s_{1n}}\Lambda_{nn} +\frac{\partial \pi^n}{\partial s_{2n}} \\
\end{bmatrix}
\\[12pt]
& F_{1r}=
\begin{bmatrix}
  R_{11} & R_{12} & \cdots & R_{1n} \\
  & R_{22} & & R_{2n} \\
 &   &  \ddots & \vdots \\
  &  &  & R_{nn}\\[10pt]
 \frac{\partial \pi^1}{\partial s_{11}}R_{11} & \frac{\partial \pi^1}{\partial s_{11}}R_{12} & \cdots & \frac{\partial \pi^1}{\partial s_{11}}R_{1n}  \\
  & \frac{\partial \pi^2}{\partial s_{12}} R_{22} & & \frac{\partial \pi^2}{\partial s_{12}} R_{2n} \\
  & & \ddots & \vdots \\
 &  &  & \frac{\partial \pi^n}{\partial s_{1n}}R_{nn}\\
\end{bmatrix}
\end{aligned}
\end{equation}

By grouping $\delta z_i$ and $\delta a_i$ from the same dimension, Eq.~\eqref{zv} is rearranged into the hierarchical combination, i.e., 
\begin{equation}
\begin{aligned}
\begin{bmatrix}
\delta \dot z_1 \\ \delta \dot a_1  \\
\delta \dot z_2 \\ \delta \dot a_2  \\
\vdots \\
\delta \dot z_n \\ \delta \dot a_n 
\end{bmatrix}
=
&\begin{bmatrix}
F_{11} & F_{12} & \cdots & F_{1n} \\
0 & F_{22} &  & F_{2n} \\
\vdots & & \ddots & \vdots \\
0 & \cdots & 0 & F_{nn}
\end{bmatrix}
\begin{bmatrix}
\delta z_1 \\ \delta a_1  \\
\delta z_2 \\ \delta a_2  \\
\vdots \\
\delta z_n \\ \delta a_n  
\end{bmatrix}
\end{aligned}
\end{equation}
where 
\begin{equation}
\begin{aligned}
     F_{ii}&=   
     \begin{bmatrix}
         \Lambda_{ii} & R_{ii}  \\
  \frac{\partial \pi^i}{\partial s_{1i}} \Lambda_{ii} + \frac{\partial \pi^i}{\partial s_{2i}} &  \frac{\partial\pi^i}{\partial s_{1i}}R_{ii}
     \end{bmatrix}
     \\[10pt]
     F_{ij}&=
     \begin{bmatrix}
    0 & R_{ij} \\
      0 &   \frac{\partial \pi^i}{\partial s_{1i}}R_{ij}
     \end{bmatrix}, \ (i<j)
\end{aligned} 
\end{equation}

\subsubsection{Linear constraints from the \textit{preservation of contraction}} Given the hierarchical combination Eq.~\eqref{zv}, if 
\begin{equation}\label{zv_ii}
\begin{bmatrix} \delta \dot z_i \\ \delta \dot a_i \end{bmatrix} = F_{ii}\begin{bmatrix} \delta z_i \\ \delta a_i \end{bmatrix}
\end{equation}
is contracting, the resulting combination is contracting. The proof of the \textit{preservation of contraction} can be found in Section 3.8.3 of \cite{lohmiller1998contraction}, Theorem 2.7 in \cite{tsukamoto2021contraction}, and in \cite{slotine2001modularity},. 

If $F_{ii}$ has real negative eigenvalues, denoted by $-\lambda_i$ where $\lambda_i>0$ ($i=1,2$), the solution of Eq.~\eqref{zv_ii} converges to zero as $t$ progresses, i.e., 
\begin{equation}
\begin{bmatrix} \delta z_i \\ \delta a_i \end{bmatrix}
= c_1\mathbf v_1 e^{-\lambda_1t} + c_2\mathbf v_2 e^{-\lambda_2t}
\end{equation}
where $c_1$ and $c_2$ are constants, $\mathbf v_1$ and $\mathbf v_2$ are the corresponding eigenvectors.

By solving the characteristic equations of $F_{ii}$, 
\begin{equation}
    s^2 -(\Lambda_{ii}+\frac{\partial \pi^i}{\partial s_{1i}}R_{ii})s - R_{ii}\frac{\partial \pi^i}{\partial s_{2i}} =0
\end{equation}
we obtain that if there exists $\alpha>0$ in a region such that $\forall \mathbf x$ and $\forall t>0$, 
\begin{equation}
\begin{aligned}
  \frac{\partial \pi^i}{\partial  s_{1i}}R_{ii} + \Lambda_{ii} <-\alpha,
\ \ \ 
  \frac{\partial \pi^i}{\partial s_{2i}} R_{ii}<-\alpha
\end{aligned}
\end{equation}
the matrix $F_{ii}$ has real negative eigenvalues, and hence Eq.~\eqref{zv_ii} is contracting, resulting in the contraction of their hierarchical combination. 

Proof is finished.

\subsection{Integration of composite variables and task space controllers for unknown environments}\label{feedback_linearization}\

\subsubsection{Reference $\mathbf p(\mathbf x, \mathbf u)$ in task space control}

Given rigid body dynamics 
\begin{equation}
 H(\mathbf q)\ddot{\mathbf q} + C(\mathbf q, \dot{\mathbf q})\dot{\mathbf q} + \mathbf g(\mathbf q) =\bm \tau \end{equation}
 where $\mathbf q$ represents the generalized coordinates, $H(\mathbf q)$ is the mass matrix, $C(\mathbf q, \dot{\mathbf q})$ represents the Coriolis terms, $\mathbf g(\mathbf q)$ is the gravity, the dynamics of the end effector can be represented as 
 \begin{equation}
 \begin{aligned}
 \mathbf x &= \mathbf h(\mathbf q)  \\
 \dot{\mathbf x} &=J(\mathbf q)\dot{\mathbf q} \\
 \ddot{\mathbf x} &=J(\mathbf q)\ddot{\mathbf q} + \dot J(\mathbf q)\dot{\mathbf q} 
 \end{aligned}
 \end{equation}

Given desired $\mathbf x_d$, one example of task space controllers without considering the null space can be represented as 
\begin{equation}\label{tau_task_space_control}
\begin{aligned}
\bm\tau= 
   &  H(\mathbf{q})J^{\dagger}\mathbf p(\mathbf x, \mathbf u)
   \\[8pt]
   &+ H(\mathbf{q})J^{\dagger}(\ddot{\mathbf x}_d -\dot J(\mathbf q)\dot{\mathbf q} )\\[8pt]
   &+C(\mathbf q, \dot{ \mathbf q})\dot{ \mathbf q} + \mathbf g(\mathbf q)
\end{aligned}
\end{equation}
the dynamic of the end effector becomes
\begin{equation}
\mathbf {\ddot x}_d - \mathbf {\ddot x} + \mathbf p(\mathbf x, \mathbf u)=0
\end{equation}

The dynamic behavior of end effectors is regulated by $\mathbf p(\mathbf x, \mathbf u)$. For example, one may design 
\begin{equation}
\mathbf p(\mathbf x, \mathbf u) = \Lambda_d^{-1}K_d(\mathbf{\dot x}_d -\mathbf{\dot x}) +  \Lambda_d^{-1}K_p(\mathbf{x}_d -\mathbf{x}) +\Lambda_d^{-1}\mathbf u 
\end{equation}
yielding the dynamics 
\begin{equation}\label{2nd_sys}
\Lambda_d\mathbf{\ddot e} + K_d\mathbf{\dot e} + K_p\mathbf e + \mathbf u=0 
\end{equation}
where $\mathbf e=\mathbf x_d -\mathbf x$.

Note that there exists different types of task space control in Section 3.2.3 in \cite{nakanishi2008operational}.

\subsubsection{Composite variables for a given $\mathbf p(\mathbf x, \mathbf u)$}

Given Eq.~\eqref{2nd_sys}, we construct a composite variable, 
\begin{equation}\label{mapping_pos}
\mathbf{y}=K_1\mathbf{e} + K_2\dot{\mathbf{e}}
\end{equation} 
where $K_1$ and $K_2$ are positive diagonal matrices to make the solution to $K_1\mathbf{e} + K_2\dot{\mathbf{e}}=0$ converging. 

We can solve for $K_1$ and $K_2$ that plugging Eq.~\eqref{mapping_pos} into Eq.~\eqref{2nd_sys} yields a dynamic model in the latent space in the form of 
\begin{equation}\label{LTI}
A\dot{\mathbf{y}} +  B\mathbf{y}+ \mathbf{u} =0
\end{equation}
where $A$ and $B$ are positive-definite diagonal matrices. 

The resulting solutions are as follows:
\begin{itemize}
    \item When $K_p=0$, we can use $\mathbf y=\mathbf{\dot e}$, yielding
    \begin{equation}\label{lambda_b_1}
\Lambda_d\dot{\mathbf{y}} + K_d\mathbf{y}+ \mathbf{u} =0
    \end{equation}
\item Otherwise, we can use $\mathbf{y}=K_1\mathbf{e} + K_2\dot{\mathbf{e}}$ where
\begin{equation}
    K_{1,i} =\frac{K_{p,i}}{\lambda_{d,i}},
 \ \ \ \ 
    K_{2,i}=  \frac{2K_{p,i}}{K_{d,i} \pm \sqrt{K_{d,i}^2-4K_{p,i}\lambda_{d,i}}}
\end{equation}
where $K_{1,i}$, $K_{2,i}$, $\lambda_{d,i}$, $K_{p,i}$, and $K_{d,i}$ denotes the $i^{th}$ diagonal element of $K_1$, $K_2$, $\Lambda_d$, $K_p$, and $K_p$ respectively. Note this requires $K_{d,i}^2-4K_{p,i}\lambda_{d,i}\geq 0$, which should be realized in the task space controller design.
The resulting dynamics becomes $A\dot{\mathbf{y}} +  B\mathbf{y}+ \mathbf{u} =0$ with
\begin{equation}\label{lambda_b_2}
A_{ii}=\frac{\lambda_{d,i}}{2K_{p,i}}(K_{d,i}\pm\sqrt{K_{d,i}^2-4K_{p,i}\lambda_{d,i}}), B_{ii}=\lambda_{d,i}
\end{equation}
\end{itemize}

\subsubsection{Stability constraints}

Provided $A$ and $B$ are positive-definite diagonal matrices, the derivatives $\frac{\partial \mathbf f}{\partial \mathbf y}= -A^{-1}B$ and $\frac{\partial \mathbf f}{\partial \mathbf u}= -A^{-1}$ are negative-definite diagonal matrices. Hence in Theorem 2, we can use identity $T_y$ and $T_a$, resulting in the inequality condition that if there exists $\alpha>0$ such that 
\begin{equation}\label{inequal_1}
\begin{aligned}
  -\frac{\partial \pi^i}{\partial  s_{1i}}A^{-1}_{ii} - A^{-1}_{ii}B_{ii} <-\alpha,
\ \ \ 
 - \frac{\partial \pi^i}{\partial s_{2i}}A^{-1}_{ii} <-\alpha
\end{aligned}
\end{equation}
the system is contracting.

Given positive-definite $A$ and $B$, the inequalities Eq.~\eqref{inequal_1} becomes positive network Jacobians.

\subsection{Robustness}\label{robustness}

\subsubsection{Robustness to unknown deterministic perturbation} 

The robustness to unknown deterministic perturbation can be found in Theorem 2.4 in \cite{tsukamoto2021contraction}, which states the robustness of a contracting system as follows: Let $\xi_0$ denote a trajectory of a contracting system $\dot{\mathbf y}= \mathbf g(\mathbf y, t)$ with the convergence rate $\beta$, and $\xi_1$ a trajectory of its perturbed system 
$\dot{\mathbf y}= \mathbf g(\mathbf y, t)+\mathbf d(\mathbf y, t)$. The distance between $\xi_0$ and $\xi_1$ exponentially converges to a bounded error ball as long as $\Theta \mathbf d$ is bounded, where $\Theta$ is the smooth coordinate transformation that constructs the contraction metric $M=\Theta^T\Theta$. Specifically, if $\exists \underline m, \bar m\in\mathbb{R}_{>0}$ and $\exists \bar d\in \mathbb R_{\geq 0}$ s.t. $\bar d = \sup_{\mathbf y, t}\lVert \mathbf d(\mathbf y,t)\rVert$ and $\underline m I \preceq M(\mathbf y,t) \preceq \bar mI$, $\forall x, t$, then we have the following relation:
\begin{equation}
\lVert \xi_1(t) -\xi_0(t) \rVert \leq \frac{V_l(0)}{\sqrt{\underline m}}e^{-\beta t} + \frac{\bar d}{\beta}\sqrt{\frac{\bar m}{\underline m}}(1-e^{-\beta t})
\end{equation}
where $V_l(0)$ denotes the path integral at time zero between $\xi_1(0)$ and $\xi_0(0)$.

Given a contracting system $ \dot{\mathbf y} = \mathbf f(\mathbf y, \mathbf u, t)$ with the convergence rate $\beta$, we consider the perturbed dynamic
\begin{equation}\label{fd}
\begin{aligned}
 \dot{\mathbf y} &= \mathbf f(\mathbf y, \mathbf u, t)  + \mathbf d(\mathbf y, \mathbf u, t) \\[6pt]
 \mathbf u &=T_a\bm\pi(\mathbf s_1, \mathbf s_2) = T_a\bm\pi(\mathbf z, \smallint \mathbf zdt)
 \end{aligned}
\end{equation}
where $\mathbf d$ represents unknown, bounded, and deterministic disturbances.

Defining $\mathbf{\bar y}=[\mathbf y^T, \mathbf u^T]^T$, 
the perturbed dynamic can be represented as 
\begin{equation}
\mathbf{\dot{\bar y}}=\mathbf{\bar f}(\mathbf{\bar y}, t) + \mathbf{\bar d} 
\end{equation}
where 
\begin{equation}
\begin{aligned}
&\mathbf {\bar f}(\mathbf{\bar y}, t) \\
=&
\begin{bmatrix}\mathbf f(\mathbf y, \mathbf u, t) \\ 
\dot T_a\bm\pi + (T_a\frac{\partial \bm\pi}{\partial \mathbf s_1}\dot T_y + T_a \frac{\partial \bm\pi}{\partial \mathbf s_2}T_y)\mathbf y + (T_a\frac{\partial \bm\pi}{\partial \mathbf s_1}T_y)f(\mathbf y, \mathbf u, t)
\end{bmatrix}
\end{aligned}
\end{equation}
and 
\begin{equation}
 \mathbf{\bar d} =
\begin{bmatrix}
 \mathbf d(\mathbf y, \mathbf u, t) \\ (T_a\frac{\partial \bm\pi}{\partial \mathbf s_1}T_y) \mathbf d(\mathbf y, \mathbf u, t)
\end{bmatrix}
\end{equation}

Therefore, applying Theorem 2.4 in \cite{tsukamoto2021contraction}, the distance between the trajectories of the contracting system and its perturbed dynamics is converging to a bounded error ball with the center at the equilibrium in the following way, 
\begin{equation}
\begin{aligned}
  \lVert \xi_1(t)-\xi_2(t) \rVert 
   \leq C_1e^{-\beta t}  + C_2\sup_{\mathbf y,\mathbf u, t} \lVert \Theta_M\mathbf{\bar d}\rVert \frac{1-e^{-\beta t}}{\beta}
\end{aligned}
\end{equation}
where $\Theta_M$ is in Eq.~\eqref{theta_m}, $C_1$ and $C_2$ are constants determined by the bounds of $M=\Theta_M^T\Theta_M$ and the initial conditions $\xi_0(0)$ and $\xi_1(0)$, provided with bounded $T_a$, $T_y$, and $\partial \bm\pi/\partial \mathbf s_1$. 

\subsubsection{Robustness to model errors} 
Let $\hat {\mathbf f}$ denote the model used to compute the constraints and $\mathbf f$ the real dynamics. Applying the stability theorem in Section.~\ref{theorem_1}, we can obtain the following result in robustness.

Given $\hat T_y$, $\hat T_a$ that are computed with $\hat{\mathbf f}$, the system remains stable at the presence of model errors, 
if there exists $\beta>0$ such that $\forall \mathbf x$, $\forall t>0$,
\begin{equation}
 (F_{1,real}^T +F_{1,real})\preceq (\hat F_1^T + \hat F_1)\prec -[\beta + max(\nu^+,0)]I
\end{equation}
where
\begin{equation}
\begin{aligned}
F_{1, real}&= 
\begin{bmatrix}
\hat T_y\frac{\partial \mathbf{f}}{\partial \mathbf{y}}\hat T_y^{-1} & \hat T_y\frac{\partial \mathbf{f}}{\partial \mathbf{u}} \hat T_a \\[8pt]
\frac{\partial \bm\pi}{\partial \mathbf{s}_1}
\hat T_y\frac{\partial \mathbf{f}}{\partial \mathbf{y}}\hat T_y^{-1} + \frac{\partial \bm\pi}{{\partial \mathbf{s}_2}} & \frac{\partial \bm\pi}{\partial \mathbf{s}_1}
\hat T_y\frac{\partial \mathbf{f}}{\partial \mathbf{u}} \hat T_a 
\end{bmatrix}
\\[8pt]
\hat F_{1}&= 
\begin{bmatrix}
\hat T_y\frac{\partial \hat{\mathbf{f}}}{\partial \mathbf{y}}\hat T_y^{-1} & \hat T_y\frac{\partial\hat{\mathbf{f}}}{\partial \mathbf{u}} \hat T_a \\[8pt]
\frac{\partial \bm\pi}{\partial \mathbf{s}_1}
\hat T_y\frac{\partial \hat {\mathbf{f}}}{\partial \mathbf{y}}\hat T_y^{-1} + \frac{\partial \bm\pi}{{\partial \mathbf{s}_2}} & \frac{\partial \bm\pi}{\partial \mathbf{s}_1}
\hat T_y\frac{\partial \hat{\mathbf{f}}}{\partial \mathbf{u}} \hat T_a 
\end{bmatrix}
\end{aligned}
\end{equation}
and $\nu^+$ is the upper bound of the eigenvalues of $F_2^T+F_2$ in Eq.~\eqref{F2}.

\subsection{Implementation in the PPO example}\label{PPO}

The dynamics of the peg-robot is 
\begin{equation}
\tau_x \dot x + x = u_x, \ \  \tau_z \dot z + z = u_z 
\end{equation}
where $\tau_x=0.0437$, $\tau_z=0.01$, $x\in[1, 3]$, and $z\in[-3, 1]$. Given the surface profile $g(x)=K_1\sin(x)+K_2\cos(x)$ with randomly sampled coefficient $K_1\in [-0.01, 0.01]$ and $K_2\in [-0.01, 0.01]$, the contact force can be approximated by 
\begin{equation}
f = K_{sur}\min{(z-g(x),\  0)} 
\end{equation}
where $K_{sur}$ represents the stiffness of the surface material, randomly sampled from $K_{sur}\in [1, 31]$. The desired $x_d$ and $f_d$ are also randomly sampled at the beginning of each trajectory.

The implementation includes (a) computation of $T_y$, $T_a$, $\Lambda_{ii}$, and $R_{ii}$ and (b) simplification into constraints on MLP weights. 

\paragraph{Computation of $T_y$, $T_a$, $\Lambda_{ii}$, and $R_{ii}$}
When the peg is in contact with the surface, the dynamic of $x$ and $f$ can be represented as 
\begin{equation}
\begin{aligned}
\dot x = & -\frac{1}{\tau_x}x + \frac{1}{\tau_x}u_x \\
\dot f = &K_{sur}(\dot z - \frac{dg}{dx}\dot x) \\
= 
&-\frac{1}{\tau_z}f + [\frac{K_{sur}}{\tau_x}\frac{dg}{dx}x -\frac{K_{sur}}{\tau_z}g(x)]  \\
&-\frac{K_{sur}}{\tau_x}\frac{dg}{dx}u_x  + \frac{K_{sur}}{\tau_z} u_z\\
\end{aligned}
\end{equation}
Defining $\mathbf x=[x, f]^T$, the nonlinear system model $\dot{\mathbf x} = \mathbf f(\mathbf x, \mathbf u, t)$ becomes 
\begin{equation}
\begin{aligned}
\dot{\mathbf x} =
\frac{d}{dt}
\begin{bmatrix}
x \\ f
\end{bmatrix}
= 
&\begin{bmatrix}
-\frac{1}{\tau_x}x
\\[7pt]
 [\frac{K_{sur}}{\tau_x}\frac{dg}{dx}x -\frac{K_{sur}}{\tau_z}g(x)] -\frac{1}{\tau_z}f 
\end{bmatrix} 
\\[8pt]
&+
\begin{bmatrix}
 \frac{1}{\tau_x} & 0 \\[7pt]
 -\frac{K_{sur}}{\tau_x}\frac{dg}{dx} &  \frac{K_{sur}}{\tau_z} 
\end{bmatrix}
\mathbf u
\end{aligned}
\end{equation}
where $g(x)$ is the surface profile. 

Instead of accurately computing the partial derivatives $\partial \mathbf f/\partial \mathbf x$ and $\partial \mathbf f/\partial \mathbf u$, we approximate the derivative with 
\begin{equation}
\begin{aligned}
\frac{\partial \mathbf f}{\partial \mathbf x} &\approx 
\begin{bmatrix}
-\frac{1}{\tau_x} & 0\\[7pt]
\frac{K_{sur}}{\tau_x}\frac{dg}{dx} -\frac{K_{sur}}{\tau_z}\frac{g(x)}{x}  &  -\frac{1}{\tau_z}
\end{bmatrix}
  \\
\frac{\partial \mathbf f}{\partial \mathbf u} &=  
\begin{bmatrix}
 \frac{1}{\tau_x} & 0 \\[7pt]
-\frac{K_{sur}}{\tau_x}\frac{dg}{dx} &  \frac{K_{sur}}{\tau_z} 
\end{bmatrix}
\end{aligned}
\end{equation}

We then compute $T_y$, $T_a$, $\Lambda_{ii}$, and $R_{ii}$ time-stepwise according to Theorem 2.

\paragraph{Simplification into constraints on MLP weights}

Because $\Lambda_{11}= -1/\tau_x$ and $\Lambda_{22}= -1/\tau_z$ are negative, the constraints become that the signs of network Jacobians are opposite to $R_{ii}$ (Eq.~\eqref{constraint_sign}). 

We use a parallel sequence of 3-layered MLPs with hyperbolic tangent function as the activation function and with positive weights. To satisfy Eq.~\eqref{constraint_sign}, a conditional multiplication is added: if $R_{ii}>0$, the output of the control policy is multiplied by -1.

%

    \begin{table}[h!]
    \caption{Hyperparameters in Peg Maze}
    \centering
    \begin{tabular}{|c|c|c|c|c|}
    \hline
     Parameter &  TD3/C-TD3 & HIRO \\
     \hline
    Gaussian noise std & HL:1.0, LL:0.2 & HL:1.0, LL:1.0  \\
     \hline
    gradient clipping  & 0.1 & 0.1  \\
    \hline
    target update period  & 100& 100 \\
    \hline
    train steps & HL:100, LL:200 & HL:100, LL:200 \\
    \hline
    \end{tabular}
    \label{tab:my_label}
\end{table}

    \begin{table}[h!]
    \caption{Peg Push Task Settings}
    \centering
    \begin{tabular}{|c|c|c|c|}
    \hline
      Geometry &  Length (y-axis) & Width (x-axis) & Depth (z-axis) \\
    \hline 
  Box   &  36 mm   &   36 mm   &  20 mm \\
     \hline
  Sliding cover &  12 mm &   36 mm  &  4 mm \\
     \hline
 Slot  &  - &   4 mm   &  7 mm\\
    \hline
Obstacle &  2 mm &    2 mm  &  2 mm\\
    \hline
 Target & 1 mm & 1 mm &  12.5* mm  \\
         \hline
   Peg   &  -  & radius=0.8 mm     & 20 mm   \\
   \hline
    \end{tabular}
    \label{tab:geometry}
\end{table}

    \begin{table}[h!]
    \caption{Parameters of TD3-based Algorithms}
    \centering
    \begin{tabular}{|c|c|}
    \hline
 critic learning rate   & 0.001  \\
     \hline
 target learning rate  & 0.0001  \\
     \hline
 adam beta1  & 0.9  \\
     \hline
      adam beta2  & 0.999  \\
     \hline
   high-level  dqda clipping & None \\
     \hline
  low-level  dqda clipping & 0.1 \\
     \hline
target q clipping & None \\
     \hline
target tau & 0.005 \\
     \hline
     target update period &  50  \\
     \hline 
action regularizer & None \\
\hline
observation normalization & None \\
\hline
reward scaling & 1\\
\hline
high-level discount factor & 0.5 \\
\hline
low-level discount factor  & 1\\
\hline 
high-level exploration noise & 0.1$\mathcal{N}(0, 0.1) $ mm\\
\hline
low-level exploration noise & $\mathcal{N}(0, 0.1)$\\
\hline 
batch size  & 128   \\
\hline
buffer size & 65536 \\
\hline
samples to buffer per update & 1024 \\
\hline
train step per update & 100 \\
\hline 
Translation Kp  & 20  \\
     \hline
     Translation Kd  & 30  \\
     \hline
   Orientation Kp  & 30  \\
     \hline
    Orientation Kd  & 40  \\
     \hline 
 high-level critic network. & (400, 300) with tanh  \\
 \hline
low-level critic network. & (400, 300) with tanh  \\
 \hline
  high-level actor network. & (400, 300) with tanh  \\
 \hline
low-level actor network. & (200, 200) with $W>0$ and tanh  \\
 \hline
    \end{tabular}
    \label{tab:para}
\end{table}

\subsection{Peg Maze Example}\label{hiro_example}
The goal is sampled between $x\in(0.41, 0.415)$, $y\in(-0.025, -0.02)$, and $z\in(0.005, 0.015)$ with the success defined as the L2 norm is less than 0.005. The robot starts almost from with the same position with small random changes in joint position.

As for the operational space controller, the force-based control is applied, that is, Section 3.3 in \cite{nakanishi2008operational}. The task space control uses $K_{p,trans}=40$,  $K_{d,trans}=20$, , $K_{p,rot}=20$, and $K_{p,rot}=10$. The joint position controller uses $K_p=1$. Because of the high-level action in joint space is very small, we used $K_d=0$ and the low-level action from HIRO is forced to have the same sign with the error for stability guarantee. For HIRO, the high-level action in joint space is within $[-0.1, 0.1]$ that is small enough to avoid the end effector to move unstably. In TD3/C-TD3, we used the task space directly as the the high-level action space, that is, $x\in(0.40, 0.43)$, $y\in(-0.03, -0.015)$, and $z\in(-0.005, 0.015)$.  

Hyperparameters are in Table~\ref{tab:my_label} and others are the defaults from HIRO \cite{nachum2018data}. Negative norm of error is used as rewards.

\subsection{Peg-Push Example}\label{HRL_example}

The geometry is listed in Table.~\ref{tab:geometry} and the depth of target refers to the distance from the target to the top surface of the box. The initial peg position is randomly sampled in a $8 \times 8 \times 8 mm^3$  volume on the right of the sliding cover. The contact parameters in Mujoco are solref = `0.01 1' and solimp=`0.9 0.95 0.001'. Control frequency is 50 Hz and the high-level action frequency is 2.5 Hz (20 environment steps). The trajectory length is set to 20 s.

The high-level reward is $-tanh(50||\mathbf e_{trans}||) - 0.1||\mathbf e_{rot}||$ where the error is with respect to the goal (the target orientation is perpendicular to the table); the low-level reward is the negative sum of the translation error norm and the rotation error norm with respect to the high-level action. 

As for the operational space control, the acceleration-based control is used, that is, Section 3.2 in \cite{nakanishi2008operational}.
The high-level action provides the 20-step increment within the range $\pm 5$ mm for translation and $\pm 0.5$ radian for orientation. The low-level action is limited to $\pm 1$. 

Hyperparameters and controller gains are listed in Table.~\ref{tab:para}. Otherwise, default parameters are used.

\balance

\end{document}